\documentclass{article}

\usepackage[preprint]{neurips_2023}

\title{Posthoc Interpretation via Quantization}

\usepackage[utf8]{inputenc} 
\usepackage[T1]{fontenc}    
\usepackage{hyperref}       
\usepackage{url}            
\usepackage{booktabs}       
\usepackage{amsfonts}       
\usepackage{nicefrac}       
\usepackage{microtype}      
\usepackage{xcolor}         

\usepackage{microtype}
\usepackage{graphicx}
\usepackage{subfigure}
\usepackage{booktabs} 
\usepackage{xparse,xfp}
\usepackage{wrapfig}

\usepackage{hyperref}
\usepackage{multirow}

\usepackage{amsmath}
\usepackage{amssymb}
\usepackage{mathtools}
\usepackage{amsthm}

\usepackage[capitalize,noabbrev]{cleveref}
\usepackage{float}
\floatstyle{plain} 
\newfloat{Code}{H}{myc}

\theoremstyle{plain}

\theoremstyle{definition}

\theoremstyle{remark}

\usepackage{tikz}
\usetikzlibrary{shapes,arrows,snakes}
\usetikzlibrary{arrows.meta,arrows}
\usepackage{hyperref}
\usepackage{listings}
\usepackage{xcolor}

\definecolor{codegreen}{rgb}{0.1, 0.5, 0.1}
\definecolor{codegray}{rgb}{0.5, 0.5, 0.5}
\definecolor{codepurple}{rgb}{0.58, 0, 0.82}
\definecolor{backcolour}{rgb}{0.95, 0.95, 0.92}

\lstdefinelanguage{PythonCustom}
{
    language=Python,
    sensitive=true,
    morekeywords={def, return},
    morecomment=[l]{\#},
    morestring=[b]",
}

\lstdefinestyle{desert}{
    backgroundcolor=\color{backcolour},
    commentstyle=\color{codegreen},
    keywordstyle=\color{magenta},
    numberstyle=\tiny\color{codegray},
    stringstyle=\color{codepurple},
    basicstyle=\ttfamily\footnotesize,
    breakatwhitespace=false,
    breaklines=true,
    captionpos=b,
    keepspaces=true,
    numbers=left,
    numbersep=5pt,
    showspaces=false,
    showstringspaces=false,
    showtabs=false,
    tabsize=2,
    moredelim=**[is][\color{red}]{@}{@},
    moredelim=**[is][\color{orange}]{&}{&},
    moredelim=**[is][\color{violet}]{~}{~}
}

\usepackage[textsize=tiny]{todonotes}

\author{%
  Francesco Paissan\footnotemark[1]~~$^{4}$, Cem Subakan\thanks{Equal Contribution}~~$^{1,2,3}$,  Mirco Ravanelli$^{2,3}$ \\
   \\
  $^1$Université Laval, $^2$Concordia University, $^3$Mila, Québec AI Institute, $^4$University of Trento
   \\
}

\begin{document}

\maketitle


\newcommand{\thetitle}{Posthoc Interpretation via Quantization}
\newcommand{\cem}[1]{\textbf{$\mathcal{CEM}$: #1}}
\newcommand{\franz}[1]{\textbf{$\mathcal{Francesco}$: #1}}

\tikzstyle{specialblock} = [draw, ultra thick, fill=blue!20, rectangle, 
    minimum height=3em, minimum width=4em]
\tikzstyle{block} = [draw, fill=lightgray, rectangle, 
    minimum height=3em, minimum width=4em]
\tikzstyle{sum} = [draw, fill=white, circle, node distance=1cm]
\tikzstyle{prod}   = [circle, minimum width=8pt, draw, inner sep=0pt, path picture={\draw (path picture bounding box.south east) -- (path picture bounding box.north west) (path picture bounding box.south west) -- (path picture bounding box.north east);}]
\tikzstyle{sumt}   = [circle, minimum width=8pt, draw, inner sep=0pt, path picture={\draw (path picture bounding box.east) -- (path picture bounding box.west) (path picture bounding box.south) -- (path picture bounding box.north);}]
\tikzstyle{input} = [coordinate]
\tikzstyle{output} = [coordinate]
\tikzstyle{pinstyle} = [pin edge={to-,thin,black}]
\tikzset{
tmp/.style  = {coordinate}, 
dot/.style = {circle, minimum size=#1,
              inner sep=0pt, outer sep=0pt},
dot/.default = 6pt 
}




\begin{abstract}


In this paper, we introduce a new approach, called \emph{Posthoc Interpretation via Quantization (PIQ)}, for interpreting decisions made by trained classifiers. Our method utilizes vector quantization to transform the representations of a classifier into a discrete, class-specific latent space. The class-specific codebooks act as a bottleneck that forces the interpreter to focus on the parts of the input data deemed relevant by the classifier for making a prediction. Our model formulation also enables learning concepts by incorporating the supervision of pretrained annotation models such as state-of-the-art image segmentation models. We evaluated our method through quantitative and qualitative studies involving black-and-white images, color images, and audio. As a result of these studies we found that PIQ generates interpretations that are more easily understood by participants to our user studies when compared to several other interpretation methods in the literature. 

\end{abstract}

\section{Introduction}

Deep neural networks have shown remarkable performance in various classification tasks, but they often remain opaque, making it hard for humans to comprehend how they make decisions. Interpretability is the ability to understand and explain a model's predictions. This desirable property is particularly valuable in areas such as healthcare, where collaboration and mutual understanding between humans and AI systems are crucial.

This paper proposes a method for interpreting neural network decisions by reconstructing relevant parts of the input data through vector quantization. This approach is a step towards achieving the ``understandability'' principle outlined in \cite{gilpin2018}, which aims to answer the question, ``\textit{Why does this particular input lead to that particular output?}''. Our goal is to provide clear, human-understandable explanations for neural network decisions, highlighting the specific parts of the input that influence the outcome.

\begin{figure}
    \centering
    \includegraphics[width=.186\textwidth]{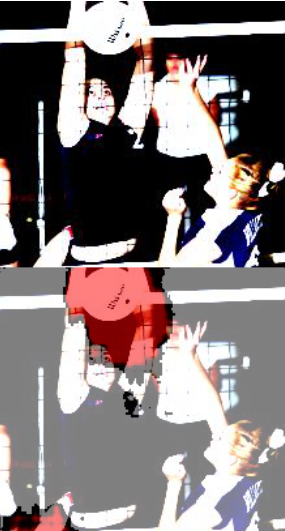}
    \includegraphics[width=.19\textwidth]{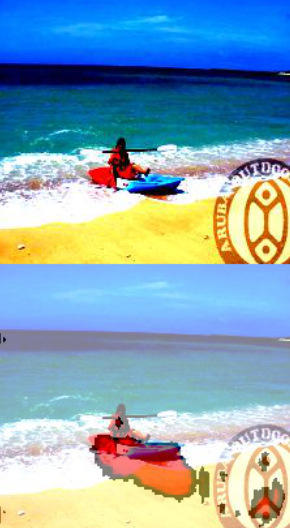}
    \includegraphics[width=.185\textwidth]{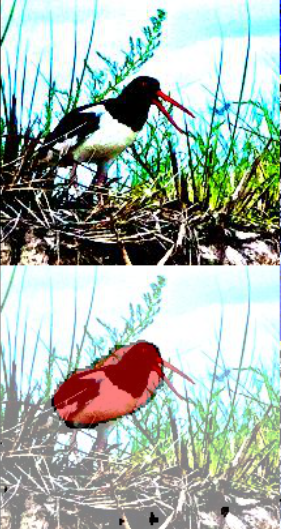}
    \includegraphics[width=.19\textwidth]{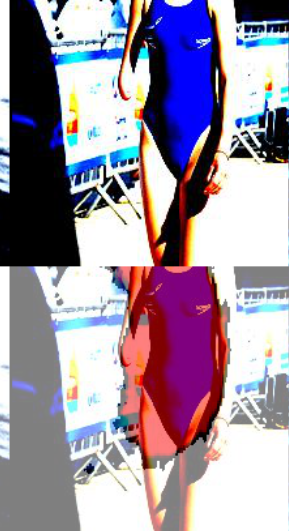}
    \includegraphics[width=.178\textwidth]{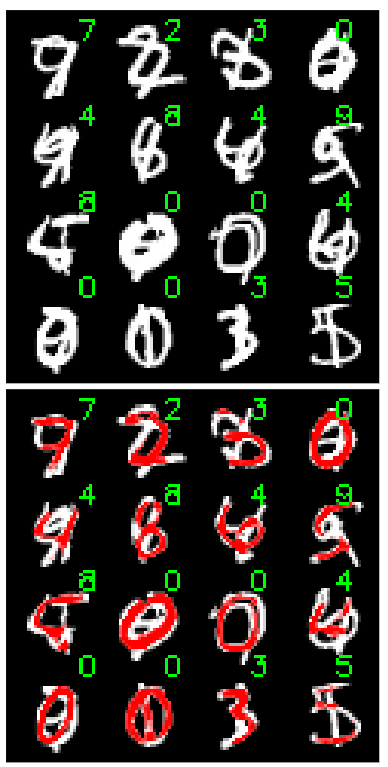}
    \caption{Showcasing the classifier interpretations generated by PIQ. (\textbf{top row}) Input images. (\textbf{bottom row}) Classifier interpretation generated by PIQ. On the first four columns we show example interpretations for the classifier decisions `volleyball', `canoe', `black-white bird' and `swimsuit'. On the last column, we show interpretations for overlapping MNIST digits. The green overlays in the top-right corner show the classifier decisions for inputs shown on the top row.}
    \vspace{-.4cm}
    \label{fig:introshowcase}
\end{figure}

In Figure \ref{fig:introshowcase}, we show several example use-cases for neural network interpretations. In the first four columns, we show the explanations provided with our method for classifications of four real-life images. We observe that the method highlights the salient objects in the image which triggers the classifier decision. Also in the last column, we show overlapping digits from the MNIST dataset \cite{lecun-mnisthandwrittendigit-2010}. 
As can be observed, it is hard to discern the dominant digit. To gain insight into how the neural network makes its decision, it would be useful to identify which parts of the image it focuses on. We show the output of our method as red overlays applied on top of the input images. We can see that the explanations provided by PIQ emphasize the parts of the input that correspond to the classifier's decisions (shown in green text). Additionally, our approach can be straightforwardly applied to audio as well, with examples available on our companion website\footnote{\url{https://piqinter.github.io/}}.


 To accomplish this, PIQ learns specific latent representations for each class. In particular, we embed the classifier's latent representations into a discrete latent space that is compartmentalized according to the classes available in the training dataset. PIQ can directly learn concepts from data modalities such as audio and black-and-white images in which saliency can be obtained by simple thresholding. PIQ can also be straightforwardly extended to more complex data such as real-life images since our framework makes it possible to extract class specific-concepts by incorporating supervision from foundational models such as the recently released Segment-Anything Model (SAM) \cite{kirillov2023segment}. 
 
 To train our interpretation module, we use the vector quantization objective, which was first introduced for Vector-Quantized VAE \cite{vqvae_vanoord}, to discretize this latent space. This discrete space acts as a bottleneck that forces the interpreter to focus on the parts of the input that are relevant to the classifier's decision. 




We present experimental results on images and audio. On images, we provide evidence on handwritten digits from the MNIST dataset \cite{lecun-mnisthandwrittendigit-2010}, clothing items from the FashionMNIST dataset \cite{xiao2017fashionmnist}, hand drawings from the Quickdraw dataset \cite{quickdraw}{, and real-world images from the ImageNet dataset \cite{Russakovsky2014ImageNetLS}.} For audio, we show results on audio clips for sound events from the ESC50 dataset \cite{piczak2015dataset}. We quantitatively evaluate our method on clean image datasets. Moreover, we provide qualitative analysis for the cases where the inputs are contaminated with samples from the same dataset (similar to the overlapping digits in Figure \ref{fig:overlapmnist}) or different datasets (as shown in Figure \ref{fig:overlapmnist}). We also perform a user study of human preferences by comparing PIQ to previous methods such as LIME \cite{LIME}, VIBI \cite{VIBI}, FLINT \cite{FLINT}, L2I \cite{parekh2022listen}{, and GradCAM \cite{Selvaraju2016GradCAMVE}}.
In summary, our contributions are the following:
\begin{itemize}
\setlength\itemsep{.003cm}
\item We introduce PIQ, a post-hoc neural network interpretation method that utilizes vector quantization to learn class-specific concepts.

\item We show that PIQ quantitatively outperforms other interpretation methods on black-and-white images. 



\item Through a series of  user studies on black-and-white images, large color images, and audio, we also show that PIQ  interpretations are preferred by humans when compared to several interpretation methods.
\end{itemize}



\subsection{Related Work}
\noindent \textbf{Concept based Posthoc-Interpretation} \\
Concept-based posthoc interpretation methods generate interpretations by defining high-level concepts. There are a variety of approaches that use concepts that are defined by a set of predefined images, such as those found in \cite{pmlr-v80-kim18d, ghorbani2019, yeh2019}. Similarly, our model learns concepts specific to each class in the latent space and stores them in the vector quantization dictionary.


Recent approaches such as listen-to-interpret (L2I) \cite{parekh2022listen} and the Framework to Learn with Interpretation (FLINT) \cite{FLINT} also aim to learn sets of features that can reconstruct the data from classifier representations. They then measure the relevance between these features and the classes to produce interpretations, with FLINT utilizing a model's output as a partial initialization for the Activation Maximization procedure \cite{AM-Mahendran2016}. However, these approaches have some limitations.
Their interpretation quality heavily relies on the relevance estimate's accuracy, which is determined by an auxiliary classifier.
 Our method is similar in that we also keep a set of features (i.e., the vector quantization dictionary), but we differ in the way we assign dictionary elements to concepts and do not require a relevance estimate, {nor} an auxiliary classifier.
 


\noindent \textbf{Other methods for Posthoc-Interpretation} \\
A widely adopted approach in the literature for creating posthoc interpretations is input attribution, as seen by methods such as GradCAM \cite{gradcam}, LIME \cite{LIME}, and other variations \cite{MONTAVON20181, lundberg2017}. These methods probe the input or intermediate representations to generate clear explanations. Other approaches exploit rule-based systems to create visual explanations, such as in the work of Ribeiro et al. \cite{ribeiro2018}. Reinforcement learning-based solutions with custom reward functions to provide text explanations, like in the research of Hendricks et al. \cite{hendricks2016}, has also been explored as well.

Another related technique is the Variational Information Bottleneck for Interpretation (VIBI) \cite{VIBI}, which uses an information bottleneck to generate an interpretation. PIQ utilizes a bottleneck representation as well. However, the way VIBI generates explanations differs from our approach as PIQ uses vector quantization and a specialized dictionary structure. We found PIQ to outperform VIBI in both quantitative and qualitative studies.



\noindent \textbf{Vector Quantized Variational Autoencoder} \\
Vector-Quantized Variational Autoencoder (VQ-VAE) \cite{vqvae_vanoord}, is an autoencoder where a bottleneck representation is vector quantized. The vector quantization enables learning discrete prior distributions over the latent distributions, which enables learning impressive generative models \cite{vqvae2}. PIQ uses the quantization in the latent bottleneck representation to define dedicated conceptual-specific codebooks for each class, and therefore is suitable to generate interpretations.  


\newcommand{\numclasses}{N_C}
\newcommand{\dict}{D}
\newcommand{\latentdim}{K}
\newcommand{\obsdim}{L}

\section{Methodology}
\label{sec:methodology}
\subsection{Overview}

{Our method, PIQ, is a posthoc interpretation method designed to generate interpretations for trained neural networks. We outline the PIQ pipeline in Figure \ref{fig:piq}.}
PIQ generates interpretations for a given classifier decision by utilizing the classifier's intermediate representation. The process starts by passing the classifier representation through an adapter layer, which is a shallow neural network that applies the first transformation. The adapted representation is vector quantized using the portion of the VQDictionary associated with the class. The decoder finally generates the interpretation mask by transforming the classifier representation using the selected dictionary items. In our experiments, we divide the VQDictionary equally among classes.

\tikzstyle{dictsmall} = [draw, thick, fill=white!10, rectangle, 
    minimum height=1.0cm, minimum width=5cm] 
    \newcommand{\xshifts}{+4.7}
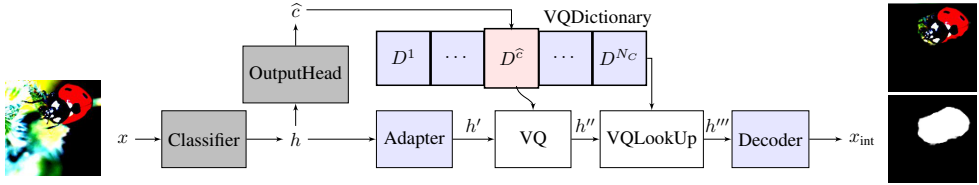
\begin{figure*}[h!]
    \centering
    \resizebox{13.2cm}{!}{
    \begin{tikzpicture}[auto, node distance=1.2cm,>=latex']
        \node [fill=none] (input) {$x$};
        \node [draw=none, fill=none, right of=input, yshift=.2cm, xshift=-2.5cm] (inppic)  { \includegraphics[scale=0.25]{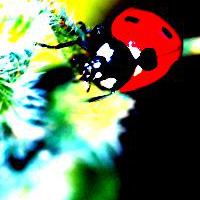}};
        \node [block, right of=input, xshift=.3cm] (cls) {Classifier}; 
        \node [right of=cls, xshift=.5cm] (h) {$h$}; 
        \node [block, above of=h] (head) {OutputHead}; 
        \node [above of=head] (chat) {$\widehat c$}; 
        \node [block, fill=blue!10, right of=h, xshift=1cm] (adap) {Adapter}; 
        \node [block, fill=white!10, right of=adap, xshift=1cm] (VQ) {VQ}; 
        \node [block, fill=white!10, right of=VQ, xshift=1cm] (StT) {VQLookUp}; 
        \node [block, fill=blue!10, right of=StT, xshift=1cm] (decoder) {Decoder}; 
        \node [right of=decoder, xshift=.5cm] (xint) {$x_\text{int}$}; 
        \node [draw=none, fill=none, right of=xint, yshift=0cm, xshift=.1cm] (outpic)  { \includegraphics[scale=0.23]{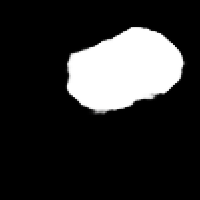}};
        \node [draw=none, fill=none, above of=outpic, yshift=.5cm] (masked)  { \includegraphics[scale=0.23]{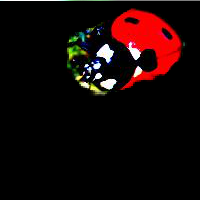}};

        \node [dictsmall, fill=blue!10, above of=VQ, yshift=0.3cm, xshift=-.4cm] (dict) {};
        \foreach \x in {1,...,4}{
            \draw [line width=0.5mm](\x + \xshifts, +1.0) -- (\x+\xshifts, 2);
        }

        \node[draw, thick, rectangle, minimum height=1.1cm, minimum width=1.0cm, fill=red!10, xshift=7.20cm, yshift=1.5cm] (red) {};
        \node [xshift=5.2cm, yshift=1.53cm] {$D^1$}; 
        \node [xshift=6.2cm, yshift=1.53cm] {$\dots$}; 
        \node [xshift=7.2cm, yshift=1.53cm] {$D^{\widehat c}$}; 
        \node [xshift=8.2cm, yshift=1.53cm] {$\dots$}; 
        \node [xshift=9.2cm, yshift=1.53cm] {$D^{\numclasses}$}; 
        \node [xshift=8.8cm, yshift=2.23cm] {VQDictionary}; 
        
        \draw [->] (input) -- (cls);
        \draw [->] (cls) -- (h);
        \draw [->] (h) -- (head);
        \draw [->] (head) -- (chat);
        \draw [->] (h) -- (adap);
        \draw [->] (adap) -- node {$h'$} (VQ);
        \draw [->] (VQ) -- node {$h''$} (StT);
        \draw [->] (StT) -- node {$h'''$} (decoder);
        \draw [->] (chat) -|  (dict);
        \draw [->] (dict) to  [out=280,in=90] (VQ);
        \draw [->] (decoder) -- (xint);
        \draw [->] (dict) -| (StT);
        
    \end{tikzpicture}
    }
    \caption{The overview of PIQ: Posthoc Interpretation via Quantization. The blue shaded boxes (VQDictionary, Decoder, and Adapter) are trained to generate interpretations for a trained classifier, represented by the gray blocks. Note the demonstration of the partition of the VQDictionary. Only the section $D_{\widehat c}$ (highlighted with red) that corresponds to class $\widehat c$ is used for the reconstruction of an input signal $x$ that is classified as $\widehat c$.  }
    \label{fig:piq}
\end{figure*}



Our model generates interpretations by breaking the Vector-Quantization dictionary into $\numclasses$ specific segments, each dedicated to a unique class ($\numclasses$ denotes the number of possible classes). This process of class-specific vector quantization creates a bottleneck in the latent space of the interpreter, allowing PIQ to reconstruct only the parts of the input that are relevant to the classifier.  The vector quantization is carried out in a learned latent space where abstract concepts are encoded and assigned to each class. We show the division of the VQDictionary items in Figure \ref{fig:piq}, within the VQDictionary block.

\begin{figure}[t]
    \centering
    \begin{tikzpicture}[auto, node distance=1.2cm,>=latex']
        \node [] (bird) {\includegraphics[scale=0.2]{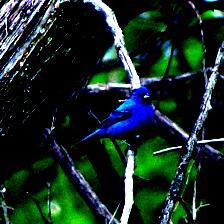}}; 
        \node [above of=bird, scale=0.8, yshift=-.3cm] {Input Image};
        \node [right of=bird, xshift=1cm] (maskbird) {\includegraphics[scale=0.2]{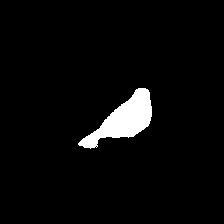}};
        \node [above of=maskbird, scale=0.8, yshift=-.3cm] {Target Mask};
        \node [left of=bird, xshift=-.9cm] (mnist1) {\includegraphics[scale=0.5]{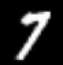}}; 
        \node [above of=mnist1, scale=.7, yshift=-.7cm] {Target Mask};
        \node [left of=mnist1, xshift=-.5cm] (mnist2) {\includegraphics[scale=0.5]{mnist_digit.png}}; 
        \node [above of=mnist2, scale=.7, yshift=-.7cm] {Input Image};

         \node [right of=maskbird, xshift=1.2cm] (spec1) {\includegraphics[scale=0.05, trim={6cm 0cm 0cm 0cm}, clip]{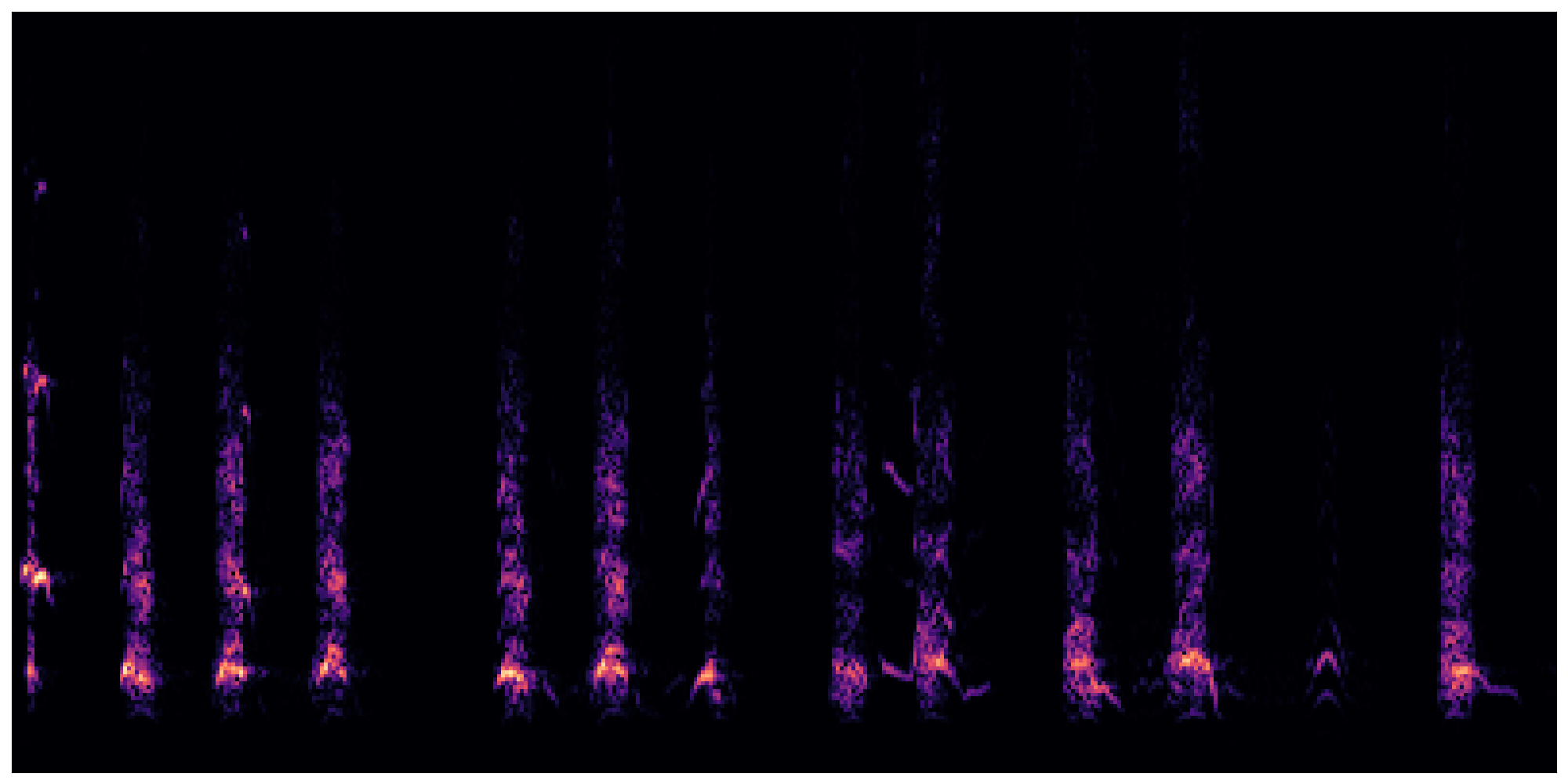}};    
         \node [above of=spec1, scale=.7, yshift=-.6cm] {Input Spectrogram};
         \node [right of=spec1, xshift=1.7cm] (spec2) {\includegraphics[scale=0.05, trim={6cm 0cm 0cm 0cm}, clip]{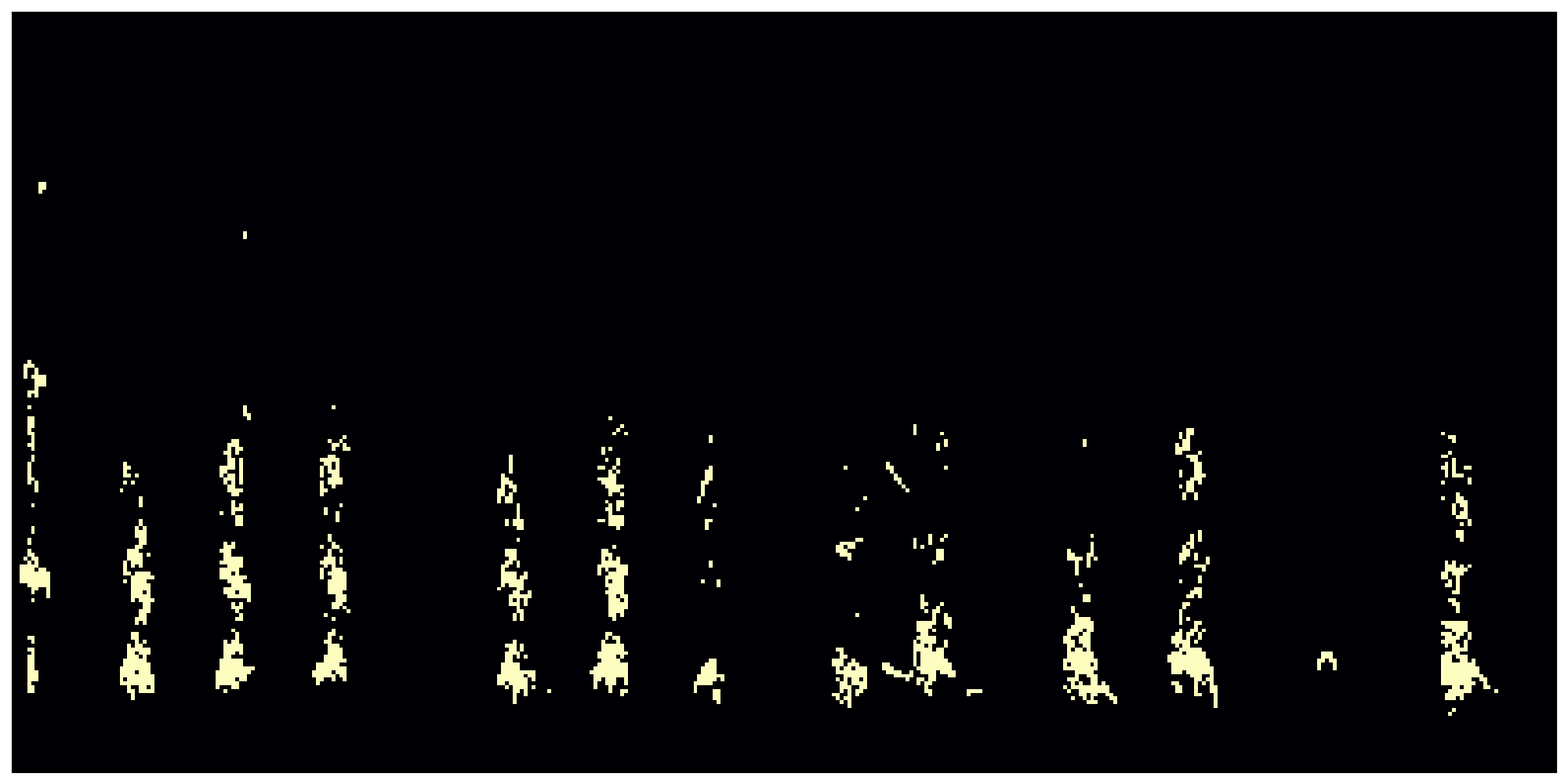}};   
         \node [above of=spec2, scale=.7, yshift=-.6cm] {Target Mask};

        \draw [->] (mnist2) -- node [xshift=-0.19cm, yshift=-.5cm, rotate=270, scale=0.7]{Identity} (mnist1);
        \draw [->] (bird) -- node [xshift=-0.19cm, yshift=-.5cm, rotate=270, scale=0.7]{SAM} (maskbird);
        \draw [->] (spec1) -- node [xshift=-0.19cm, yshift=-.55cm, rotate=270, scale=0.7]{Threshold} (spec2);
        
    \end{tikzpicture}

    \caption{Obtaining the training target masks for different data modalities. \textbf{(left)} The black-white images do not require a pre-processing step to obtain target masks. \textbf{(middle)} For real-world images, we use a segmentation model to obtain target masks during training. During inference our interpretation method works on its own. \textbf{(right)} For audio, we simply threshold the input spectrogram to obtain a binary target mask for training. }
    \vspace{-.5cm}
    \label{fig:targets}
\end{figure}

We would like to emphasize that our interpreter is trained on target interpretation masks obtained from the same training set that is used to train the classifier. Note that we do {not} train on synthetically created mixtures. 
We train PIQ to predict binary interpretation masks that highlight a specific class in the input image. For black-and-white images such as MNIST and Quickdraw images the training target masks are given by the training data itself. For audio, we simply threshold the magnitude spectra to obtain the training target masks from clean audio. {For complex images such as the ones from the ImageNet dataset, we use a foundational image segmentation model, SAM \cite{kirillov2023segment}, to obtain the training target masks. We summarize the way we obtain the training target masks in Figure \ref{fig:targets}.}

 We want to emphasize that PIQ is {not} solving a segmentation task, but rather learns to generate the interpretation mask starting from the classifier representations, via associating concepts from the VQdictionary. The details on how we use SAM to obtain the training target masks is described in Appendix \ref{app:imagenet}. We also would like to note that PIQ is able to generate interpretations for multi-label classifiers, and we provide preliminary results in Appendix \ref{app:multilabel}.

\subsection{Vector Quantization and Details on Target Data for Training PIQ}
\label{sec:vq}

The vector quantization that we use in this paper takes in a continuous representation $h \in \mathbb R^{\latentdim\times H \times W}$, (where $H$ and $W$ denote height and width of the latent representation) and assigns it to the set of closest vectors in a dictionary $D \in \mathbb R^{\latentdim\times |D|}$ that consists of $|D|$ vectors of dimension $\latentdim$. In our method, the classifier representation $h \in \mathbb R^{\latentdim \times H \times W}$, first goes through an adapter layer and we obtain $h' \in \mathbb R^{\latentdim \times H \times W}$. The quantization process is described by the following equation: 
\begin{align}
    h_{i, j}^{''} = \arg \min_k \| h'_{i, j} - D_k^{\widehat c} \|,
\end{align}
where we quantize the classifier representation $h'$ by finding the closest vector in the dictionary $D^{\widehat c}$ related to class $\widehat c \in {1, \dots \numclasses}$, for each vector $(i, j)$ in the latent representation $h'_{i,j} \in \mathbb R^\latentdim$. This results in the discretized latent representation $h''\in\mathbb Z^{H \times W}$, (which forms a grid of shape $H\times W$). By using a look-up operation, 
the discretized latent representation is then used to select the corresponding dictionary item from the dictionary, resulting in $h''':=D^{\widehat c}_{h''}$. Finally, to obtain the model output $x_\text{int} \in \mathbb R^{\obsdim}$, $h'''_{i, j}$ is passed through a decoder, yielding $x_\text{int} = \text{Decoder}(h''')$. To train the proposed posthoc interpretation model, we use the training objective defined in the original VQ-VAE paper \cite{vqvae_vanoord}, such that the training loss $\mathcal L$ is defined as,
\begin{align}
    \mathcal L = d(x_\text{int}\| x_\text{target}) + \| h' - \text{sg}(h''') \|_2^2 + \| \text{sg}(h') - h''' \|_2^2,  \label{eq:trainingloss}
\end{align}
where $d(x_\text{int}\| x_\text{target})$, denotes the reconstruction error between the estimated interpretation mask and the training target mask $x_\text{target}$, and $\text{sg}(\cdot)$ denotes the stop gradient operation. For the reconstruction loss $d(x_\text{int}\|x_\text{target})$, we use a binary loss such as negative Bernoulli likelihood for the black-white images or Dice Loss, commonly used for segmentation \cite{sudre2017generalised}, for ImageNet images. 


\section{Experiments} 

\subsection{Datasets and Model Details for Images}
\label{sec:datasetsandmodelingimages}



We evaluated PIQ both qualitatively and quantitatively on three black-and-white image datasets: MNIST \cite{lecun-mnisthandwrittendigit-2010}, FashionMNIST \cite{xiao2017fashionmnist}, and Quickdraw \cite{quickdraw}. For the Quickdraw dataset, we used a subset containing the ten classes used to evaluate FLINT \cite{FLINT}. Moreover, we qualitatively evaluated PIQ on a subset of the ImageNet dataset  \cite{Russakovsky2014ImageNetLS}, composed of the classes \emph{`indigo bunting', `oyster-catcher', `ladybug', `bathing cap', `canoe', `maillot', `mortarboard', `paddle', `steel drum', `volleyball'.} We limited the dataset 10 classes in order to be able to qualitatively evaluate the interpretation quality on all classes with a user study. 


We employed the same classifier architecture for MNIST, FashionMNIST, and Quickdraw. Specifically, we used a convolutional neural network with two convolutional blocks followed by max-pooling and a linear classifier at the end. The classification performance on MNIST, FashionMNIST, and Quickdraw datasets were $99.5\%$, $92.5\%$, and $87.0\%$, respectively. For more information on the classifier and the interpreter architecture, please refer to Appendix \ref{app:design}.
For the subset of the ImageNet dataset instead, we finetuned a ResNet-50 \cite{He2015DeepRL}, achieving a test accuracy of $88.2\%$. In this case, the interpreter decoder resembles the architecture of a VQ-VAE2 \cite{Razavi2019GeneratingDH}, with class partitioning described in Section \ref{sec:methodology}, applied to the output of the second and last convolutional stage. The two codebooks have $4096$ vectors of $2048$ entries each, uniformly distributed over classes. The output of the interpreter is a binary mask that we show on top of the original image (e.g. as in Figure \ref{fig:introshowcase}). We provide more details on this in Appendix \ref{app:imagenet}.

For the baselines, we used the original implementations of \href{https://github.com/jayneelparekh/flint}{FLINT}, \href{  https://github.com/marcotcr/lime}{LIME}, and \href{https://github.com/jacobgil/pytorch-grad-cam}{GradCAM}, which can be found on the respective GitHub repositories. For VIBI, we used a \href{https://github.com/willisk/VIBI}{recent GitHub repository}. For L2I, we used our own implementation and adapted the method to work on images as well. Additional information on the L2I implementation for images can be found in the Appendix \ref{sec:l2iimpl}. The implementation of PIQ can be found in the supplementary material. 



\subsection{Quantitative Evaluation on Images}
\label{sec:quanteval}

\begin{table*}[t]
\caption{Quantitative evaluation of interpretation quality on image datasets MNIST and FMNIST}
\label{table:quantimages}
\vskip 0.15in
\begin{center}

\resizebox{13.9cm}{!}{
\begin{tabular}{l|ccc|ccc}
\toprule
 \textbf{Dataset}  & \multicolumn{3}{c|}{MNIST} & \multicolumn{3}{c}{FashionMNIST}  \\
 \midrule
\textbf{Metric} & Fidelity-In ($\uparrow$) & Faithfulness ($\uparrow$) & FID ($\downarrow$) & Fidelity-In ($\uparrow$) & Faithfulness ($\uparrow$) & FID ($\downarrow$) \\
\midrule
 PIQ (ours) & \textbf{98.03 $\pm$ 0.05} & \textbf{0.588 $\pm$ 0.00021} & \textbf{0.029 $\pm$ 0.0004}  & 
 \textbf{81.3 $\pm$ 0.2} & \textbf{0.773 $\pm$ 0.004} &  \textbf{0.030 $\pm$ 0.0004} \\

 VIBI & 73.90 $\pm$ 16.08 & 0.369 $\pm$ 0.002 & 0.710 $\pm$ 0.962 & 42.4 $\pm$ 17.8 & 0.578 $\pm$ 0.073 & 0.395 $\pm$ 0.104 \\
 
 L2I & 96.56  $\pm$ 2.66 & 0.453 $\pm$ 0.002 & 0.160 $\pm$ 0.010 & 68.3 $\pm$ 1.5
 & 0.343 $\pm$ 0.011 & 0.188 $\pm$ 0.011
 \\
 GradCAM & 23.94 $\pm$ 0.5 & 0.0464 $\pm$ 0.001 & 0.1988 $\pm$ 0.002 & 22.89 $\pm$ 0.1 & 0.058 $\pm$ 0.003 & 0.2568 $\pm$ 0.002 \\
 FLINT & 10.9 & 0.361 & 0.677 & 15.37 & -0.097 & 0.482 \\
\bottomrule
\end{tabular}
}
\newline

\end{center}
\vspace{-.5cm}
\end{table*}

\textbf{Metrics} 

To evaluate the generated interpretations quantitatively, we use three metrics. The first one is the fidelity-to-input, which is proposed in this paper for the first time. The second metric is Fréchet-Inception-Distance (FID) \cite{heusel2017gans}, which has been used to to asses the quality of the generative models. Lastly, we use faithfulness \cite{alvarez2018} as our third metric. We define the metric of fidelity-to-input as the percentage agreement between the classifier's predictions for the original input and the interpretation. Mathematically, we express the fidelity-to-input (FID-I) as: \begin{align}
    \text{FID-I} = \frac{1}{N} \sum_{n=1}^N \left [\arg \max_c f_c(x_n) = \arg \max_c f_c(x_{\text{int}, n}) \right ], 
\end{align}
where $f_c(\cdot)$ is the classifier's output probability for class $c$, and $[\cdot]$ is the Iverson bracket which is 1 if the statement is true, and 0 if it is false. $x_n$ is the $n$'th data item, and $x_{\text{int}, n}$ is the interpretation that corresponds to the same input. This metric aims to measure how aligned the generated interpretations are to the original input in terms of the class predicted by the classifier. Ideally, the produced interpretation should not change the original classifier's decision. For example, the interpretation of a handwritten digit should not be classified as another digit.



As we mentioned in the introduction, with PIQ, we aim to generate interpretations that humans understand. Therefore, we want interpretations that are easy to associate with the original data distribution in the input space (pixel space for images). For this reason, we propose to use the Frechet-Inception Distance (FID) \cite{heusel2017gans} between the produced interpretations and the input data as an additional metric to describe the quality of interpretations. FID is a commonly used distance to measure the deviation between the distribution of the data generated by a generative model and the original data distribution. In this work, we use the original FID definition, and we extract image embeddings using an Inceptionv3 \cite{Szegedy2014GoingDW} network trained on ImageNet \cite{Russakovsky2014ImageNetLS} and compute the Fréchet Distance on the two Gaussian distributions estimated using the embeddings.

Finally, we also measure the faithfulness of the interpretations. The faithfulness metric aims to measure the importance of the interpretation to the classifier decision. By following the way L2I \cite{parekh2022listen} defines this metric, we calculate the faithfulness as,
\begin{align}
    \text{Faithfulness} = f_{\widehat c} (x) - f_{\widehat c}(x - x_\text{int}), \label{eq:faithfulness}
\end{align}
where $f_{\hat{c}}(x)$ denotes output probability for the class that corresponds to the classifier decision $\widehat c$. For example, on the overlapping digit example showcased in the introduction, if the interpretation $x_\text{int}$ recovers the original digit perfectly, subtracting it from the input data $x$ would result in a low probability in the second term of the faithfulness definition given in equation \eqref{eq:faithfulness}.  
\begin{wraptable}{l}{0.5\textwidth}
\caption{Quantitative evaluation quality on the Quickdraw Dataset}
\label{table:quickdraw}
\resizebox{0.5\textwidth}{!}{
\begin{tabular}{l|ccc}
\toprule

 \textbf{Dataset} & \multicolumn{3}{c}{Quickdraw}\\
 \midrule
\textbf{Metric} & Fidelity-In ($\uparrow$) & Faithfulness ($\uparrow$) & FID ($\downarrow$)\\
\midrule
 PIQ (ours) & \textbf{60.89 $\pm$ 0.60}& \textbf{0.675 $\pm$ 0.005} & \textbf{0.034 $\pm$ 0.0001}\\
 VIBI & 26.36 $\pm$ 3.01& 0.341 $\pm$ 0.031 &  0.388 $\pm$ 0.032\\
 L2I & 25.97 $\pm$ 0.82& 0.340 $\pm$ 0.031& 0.397 $\pm$ 0.020\\
 GradCAM & 11.32 $\pm$ 0.6 & 0.1681 $\pm$ 0.017 & 0.1882 $\pm$ 0.035 \\
 FLINT & 15.62 & -0.057 & 0.672 \\
\bottomrule 
\end{tabular}
}
\end{wraptable}

\textbf{Quantitative Performance Evaluation}

In Table \ref{table:quantimages}, we compare the quantitative metrics defined above on the three black-and-white image datasets mentioned in Section \ref{sec:datasetsandmodelingimages}. We compare PIQ with several prominent posthoc interpretation algorithms, which include FLINT \cite{FLINT}, VIBI \cite{VIBI}, GradCAM \cite{gradcam}, and Listen-to-Interpret (L2I) \cite{parekh2022listen}. We train and evaluate all the methods on clean data from their respective train and test sets. To account for training variability, we perform three runs for all methods except for FLINT, as we found its performance to be consistently worse than other methods).  
We found that PIQ outperforms the other methods in terms of FID-I, faithfulness, and FID. Furthermore, our results indicate that PIQ generates interpretations that are more closely aligned with the original data distribution, as evidenced by its lower Frechet Inception Distance (FID) values and higher fidelity-to-input (FID-I) scores. Overall, PIQ demonstrates superior performance in generating human-understandable interpretations.



\subsection{Qualitative Evaluation on Images}
\label{sec:images-qualitative} 
\textbf{Experiment description} 

\begin{figure*}[h!]
    \centering
    \includegraphics[width=0.16\textwidth]{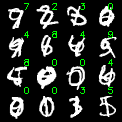}
     \includegraphics[width=0.16\textwidth]{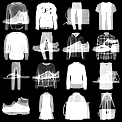}
     \includegraphics[width=0.16\textwidth]{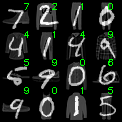}
     \includegraphics[width=0.16\textwidth]{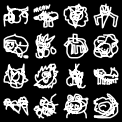}
    \includegraphics[width=0.16\textwidth]{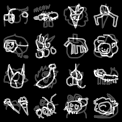}
    
    \includegraphics[width=0.16\textwidth]{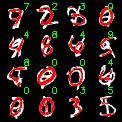}
    \includegraphics[width=0.16\textwidth]{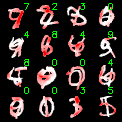}
    \includegraphics[width=0.16\textwidth]{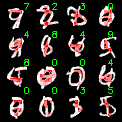}
    \includegraphics[width=0.16\textwidth]{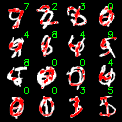}
    \includegraphics[width=0.16\textwidth]{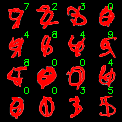}
    \includegraphics[width=0.16\textwidth]{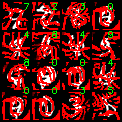}

    \includegraphics[width=0.16\textwidth]{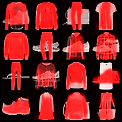}
    \includegraphics[width=0.16\textwidth]{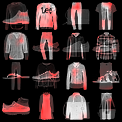}
    \includegraphics[width=0.16\textwidth]{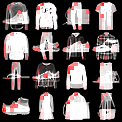}
    \includegraphics[width=0.16\textwidth]{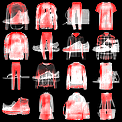}
    \includegraphics[width=0.16\textwidth]{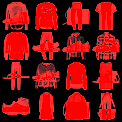}
    \includegraphics[width=0.16\textwidth]{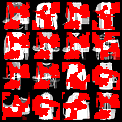}

    \includegraphics[width=0.16\textwidth]{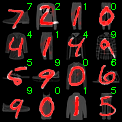}
    \includegraphics[width=0.16\textwidth]{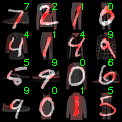}
    \includegraphics[width=0.16\textwidth]{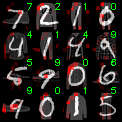}
    \includegraphics[width=0.16\textwidth]{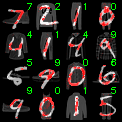}
    \includegraphics[width=0.16\textwidth]{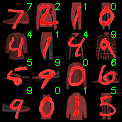}
    \includegraphics[width=0.16\textwidth]{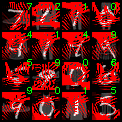}

    \includegraphics[width=0.16\textwidth]{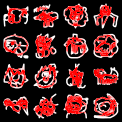}
    \includegraphics[width=0.16\textwidth]{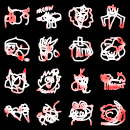}
    \includegraphics[width=0.16\textwidth]{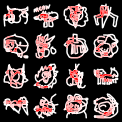}
    \includegraphics[width=0.16\textwidth]{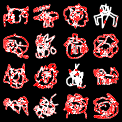}
    \includegraphics[width=0.16\textwidth]{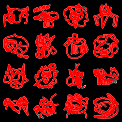}
    \includegraphics[width=0.16\textwidth]{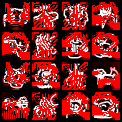}

    \includegraphics[width=0.16\textwidth]{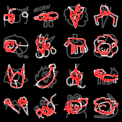}
    \includegraphics[width=0.16\textwidth]{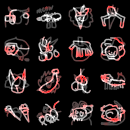}
    \includegraphics[width=0.16\textwidth]{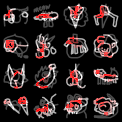}
    \includegraphics[width=0.16\textwidth]{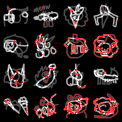}
    \includegraphics[width=0.16\textwidth]{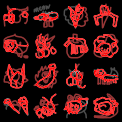}
    \includegraphics[width=0.16\textwidth]{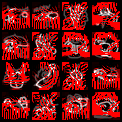}

    \caption{{Comparing interpretation methods on overlapped data. The interpretations are highlighted in red overlays. The top row shows the network's input. The second row from the top shows MNIST digits, with classifier decisions indicated on the top right corner of each digit. The third row shows overlapping FashionMNIST data items, the fourth row shows MNIST digits with FashionMNIST backgrounds, and the fifth and sixth rows show overlapping Quickdraw drawings with different weights. From left to right, interpretations are generated by PIQ, GradCAM, VIBI, L2I, LIME, FLINT,  respectively.}}
    \vspace{-.8cm}
    \label{fig:overlapmnist}
\end{figure*}

To evaluate the effectiveness of our method in handling challenging data, we performed tests on contaminated inputs. We compared various methods for generating contaminated data, specifically: \textbf{(Case1)} Overlapping Handwritten digits from the MNIST dataset \cite{lecun-mnisthandwrittendigit-2010}, \textbf{(Case2)} Overlapping Clothing items from the FashionMNIST dataset \cite{xiao2017fashionmnist}, \textbf{(Case3)} Handwritten digits with background with samples from the FashionMNIST dataset, 
\textbf{(Case4)} Overlapping Handdrawings from the Quickdraw dataset \cite{quickdraw} We have two versions where i) We overlap the images with equal weights (v1) ii) We overlap the images with weights 0.7 and 0.3 (v2).



In Figure \ref{fig:overlapmnist}, we present the interpretations generated by PIQ, VIBI, L2I, LIME, and FLINT on the challenging data setups outlined above. It's worth mentioning that the classifier predictions for cases 2, 4-i, and 4-ii can be found in Appendix \ref{sec:extrapiqsamples}. 


As the low FID values in Table \ref{table:quantimages} suggest, PIQ preserves the distribution of the handwritten digits much better than the other algorithms. {Interpretations generated by GradCAM rarely look like the original digit, as also quantitatively evidenced in Table \ref{table:quantimages} by the high FID values.} While VIBI sometimes generates interpretations that resemble digits, they often deviate from the classifier's decision, as indicated by the green indicators on the top right corner. L2I generally produces better interpretations than VIBI, but still does not attain the level of distribution preservation achieved by PIQ. LIME simply reproduces the input mixture without altering it, while FLINT's generated interpretations, even though they may contain the original digits, do not meet the criterion of understandability. 


We observe a similar behavior on the overlapping FashionMNIST items (second row of Figure \ref{fig:overlapmnist}), and MNIST digits with Fashion MNIST background (third row of Figure \ref{fig:overlapmnist}) as well. PIQ obtains interpretations that remain aligned with the classifier decision, that are easy to understand, and remain loyal to the original data distribution. Finally, on overlapping Quickdraw drawings (shown in the fourth row for the equal weight mixing case, and the fifth row for the case with weights 0.7 and 0.3 in Figure \ref{fig:overlapmnist}), we see that especially on the equal weight mixing case, the methods mostly fail to produce meaningful explanations as the mixtures are challenging. {LIME interpretations are understandable, but do not highlight any portion of the image. Therefore, LIME does not give any intuition about which part of the input contributes most to the classification.} We however observe that PIQ produces explanations that generally correlate well with the classification decisions of the classifier. Note that we provide the list of classifier decisions in $4\times4$ grid format in Appendix \ref{sec:extrapiqsamples}.   


\textbf{User Study} \\
To measure human preference towards the different interpretation methods, we performed a user study, in which we compared several different interpretation methods. Note that we have not included GradCAM in this user study as it performed worse on the quantitative metrics.



For each overlapping data case described above, we asked the participants to rate the quality of the interpretations with a score between 1 (bad) and 5 (excellent). For each case, we showed each participant 16 different images presented in a $4\times 4$ format (similar to the images in Figure \ref{fig:overlapmnist} - For case-1 we studied two batches). We first showed the participants the overlapping inputs that were given to the classifier, and then we followed up with the interpretations obtained with PIQ, VIBI, FLINT, L2I, and LIME (presented in random order). For the studies corresponding to cases 1, 2, 3, 4 we had 23, 16, 22, and 20 participants respectively. 


Table \ref{table:mosimages} displays the mean opinion scores for different interpretations using various approaches in all 4 cases. We can see that the interpretations produced by PIQ are consistently preferred by participants. In the overlapping MNIST case (case-1), there was no close contender. In case-2 (Fashion-MNIST mixtures), L2I was the second-best method in terms of participant preference. However, it's worth noting that PIQ received a score of 5 (excellent) from 15 participants, while only receiving 3 from one participant. In cases 3 and 4, LIME was the closest contender, as their interpretations tend to closely resemble the input image. However, LIME was less preferred in balanced mixtures of case 1 and 2, and more preferred in imbalanced mixtures of case 3 and 4-ii. It's also worth noting that for case 4-i and 4-ii, the study was limited to PIQ and LIME as these two methods seemed to produce the best results as seen in Figure \ref{fig:overlapmnist}.

\begin{table}[t]
\caption{Subjective evaluation of interpretation quality on overlapping black-and-white images}
\vspace{-.2cm}
\label{table:mosimages}
\begin{center}
\begin{small}
\begin{sc}
\resizebox{0.99\textwidth}{!}{
\begin{tabular}{c|c|c|c|c|c|c}
\toprule
Method & \parbox[c]{1cm}{MNIST\\(Case1)} & \parbox[c]{1.5cm}{MNIST B2\\(Case1)} & \parbox[c]{2.1cm}{FMNIST-MIX\\(Case2)} & \parbox[c]{1.8cm}{MNIST+FMN\\(Case3)} & \parbox[c]{1.7cm}{Quickdraw1\\(Case4-i)} & \parbox[c]{1.7cm}{Quickdraw2\\(Case4-ii)}  \\
\midrule
   PIQ (ours) & \textbf{4.04 $\pm$ 0.48} &  \textbf{3.95 $\pm$ 0.72}  & \textbf{4.87 $\pm$ 0.50} & \textbf{4.78 $\pm$ 0.43} &\textbf{2.6 $\pm$ 1.67} & \textbf{3.55 $\pm$ 1.0} \\
   VIBI &  1.77 $\pm$ 0.68 & 1.86 $\pm$ 0.71 & 1.37 $\pm$ 0.50 & 1.14 $\pm$ 0.47 & - & - \\
 L2I &  2.4 $\pm$ 0.66  & 1.86 $\pm$ 0.56 & 3.18 $\pm$ 0.91 & 2.18 $\pm$ 0.96 & - & -\\
  FLINT & 1 $\pm$ 0 & 1.04 $\pm$ 0.21 & 1.12 $\pm$ 0.50 & 1.09 $\pm$ 0.47 & - & - \\
  LIME  & 2 $\pm$ 1.34 & 2.13 $\pm$ 1.21 & 1.37 $\pm$ 0.89 & 3.23 $\pm$ 0.72 &2.35 $\pm$ 1.46 & 3 $\pm$ 1.38 \\
\bottomrule
\end{tabular}
}
\end{sc}
\end{small}
\end{center}
\vspace{-.7cm}
\end{table}

\vspace{-.2cm}
\subsection{Qualitative Interpretation Study on ImageNet Images}
{We have also conducted a user study to evaluate the perceived quality of the interpretations produced by PIQ on real-world images from the ImageNet dataset, and to compare these with the interpretations produced by GradCAM \cite{gradcam}.} In this study we have presented the original images, and the interpretations superposed (as shown in Figure \ref{fig:introshowcase}) on the original images, and asked the users to give their opinion on a scale from one to five, for each method. We show the exact prompt in the supplemental material. {Overall, 23 participants took part in this user study.}
\begin{wrapfigure}{l}{0.54\textwidth}
    \centering
    \includegraphics[width=.50\textwidth, trim=2.0cm .5cm 2.5cm 0cm,clip]{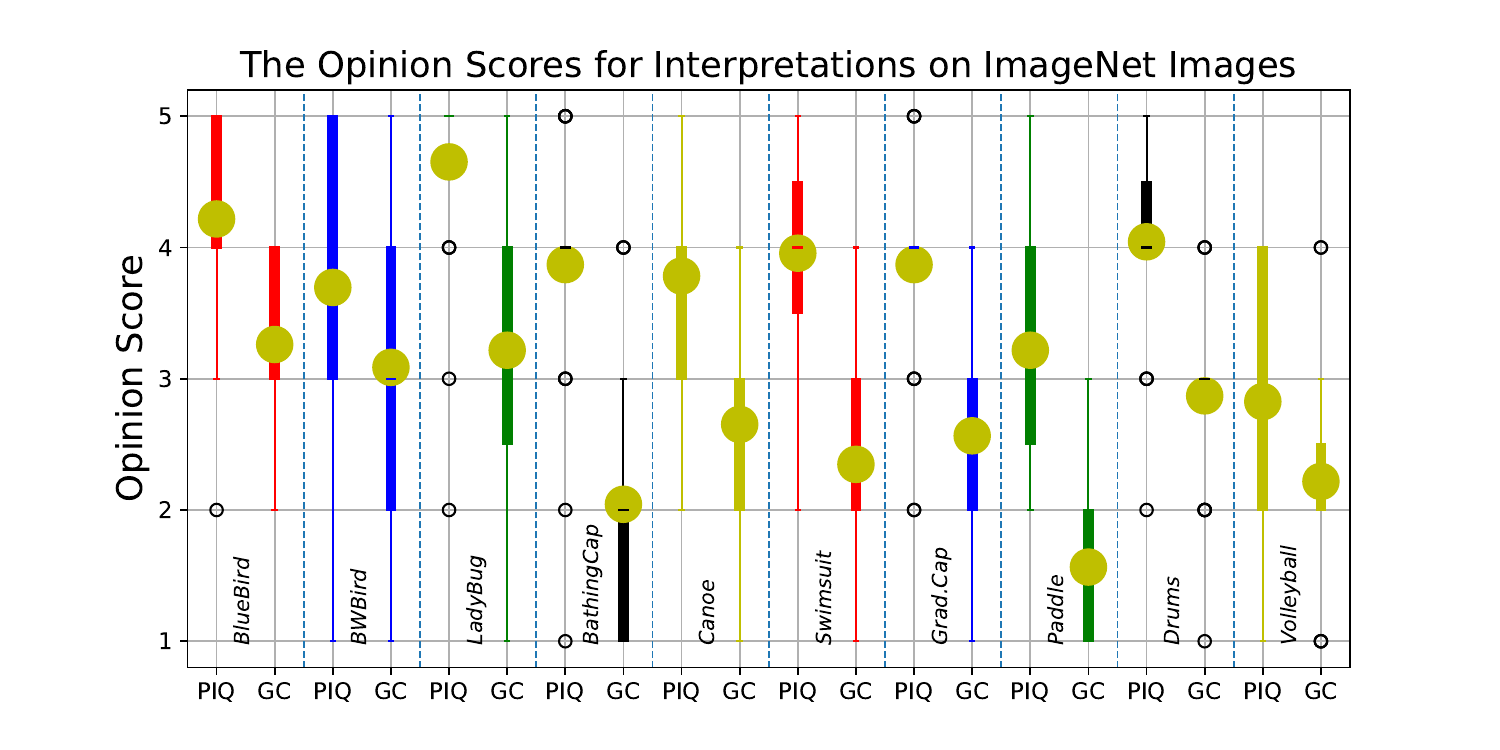}
    \vspace{-.4cm}
    \caption{Average opinion scores obtained with PIQ and GradCAM on ImageNet images. Each boxplot corresponds to the average opinion score obtained with first PIQ, then GradCAM on a series of classes. The classes are indicated on the bottom of the plot. }
    \label{fig:imagenetmos}
    \vspace{-.6cm}
\end{wrapfigure}

We summarize the result of the study in Figure \ref{fig:imagenetmos}, where for each class we show the distribution of the opinion scores with a boxplot. We see that for each class the mean-opinion-score (MOS) shown with the yellow circles on top of the box plots, is better for PIQ compared to GradCAM. We also note that the MOS for PIQ is $3.63$ and for GradCAM is $2.53$. 

We also conducted a model simulation study as proposed in \cite{Liang2022MultiVizTV}. The objective of model simulation is to measure the accuracy of the human participants on classifying the interpretations generated by PIQ and GradCAM. We asked the users to classify an example from each class for each methods. We measured an accuracy of $86.1\%$, $88.0\%$ for PIQ, GradCAM respectively. We note that in general because the PIQ interpretations are more specific, it is harder for the users to extract information from the context (PIQ removes the background more than GradCAM). But overall, we see that even though the PIQ interpretations are more specific, the users were able to obtain a similar accuracy. We show the accuracy distribution of each method along with our user prompts in Appendix \ref{app:imagenet-users}.
\vspace{-.3cm}
\subsection{Qualitative Interpretation Study on Audio}
\vspace{-.1cm}
\textbf{Dataset and Modeling Details} 
We test the interpretations produced by PIQ on the ESC-50 dataset \cite{Piczak2015ESCDF}, which consists of 2000, 5 seconds-long clips of 50 different classes of sound events. Example sound events in the dataset include `cat', `dog', `baby cry', `church-bells', and so on. As a classifier, we utilized a convolutional network consisting of four strided 2D-convolutional layers with a downsampling factor of 2. The classifier operates in the log-spectrogram domain and achieved 75\% classification accuracy. We provide the further details regarding the classifier, dataset and the interpreter architecture in the supplemental material.  
\vspace{-.0cm}

\textbf{Qualitative Evaluation and User Study} 
\begin{figure*}[t]
    \centering
    \includegraphics[width=0.40\textwidth, trim=2cm 0.5cm 1cm 0cm,clip]{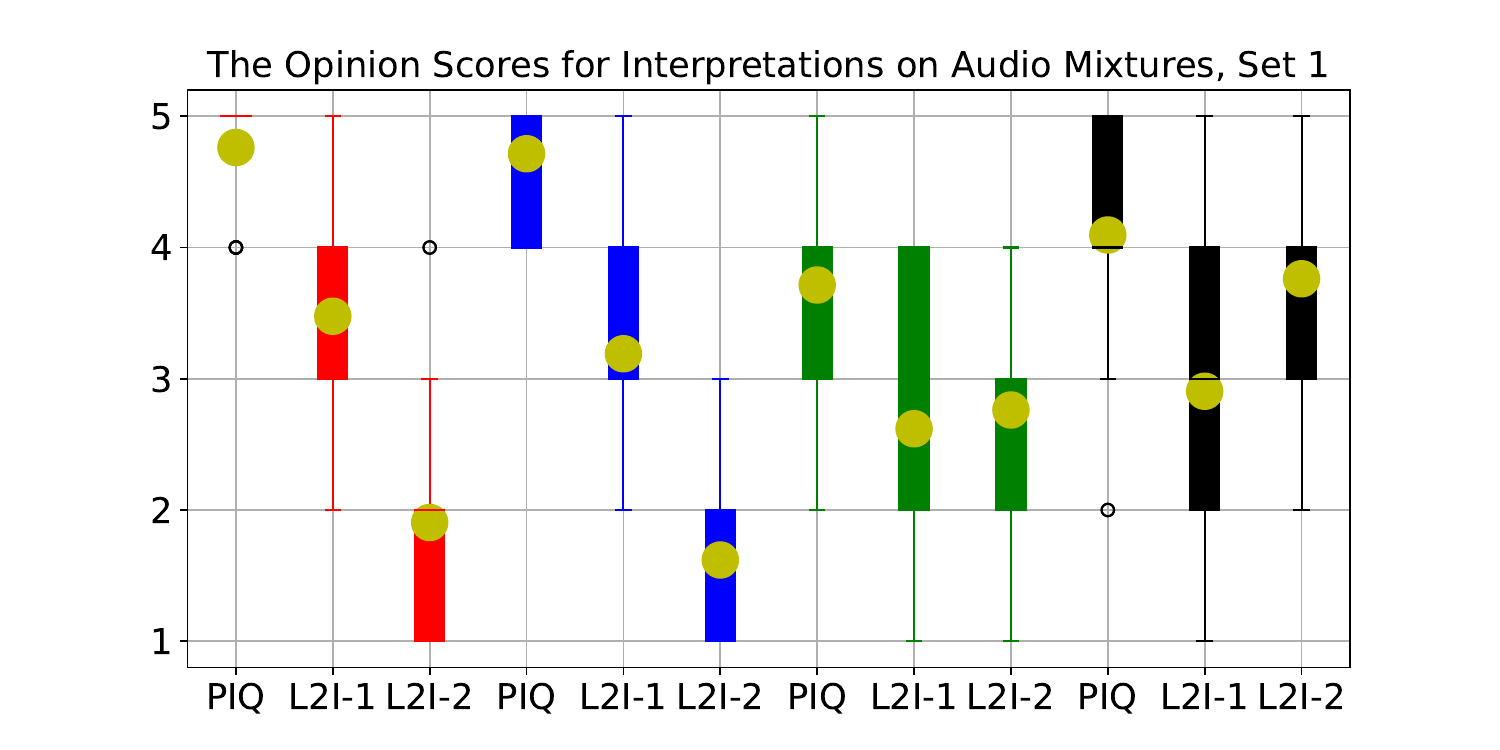}
    \includegraphics[width=0.40\textwidth, trim=2cm 0.5cm 1cm 0cm,clip]{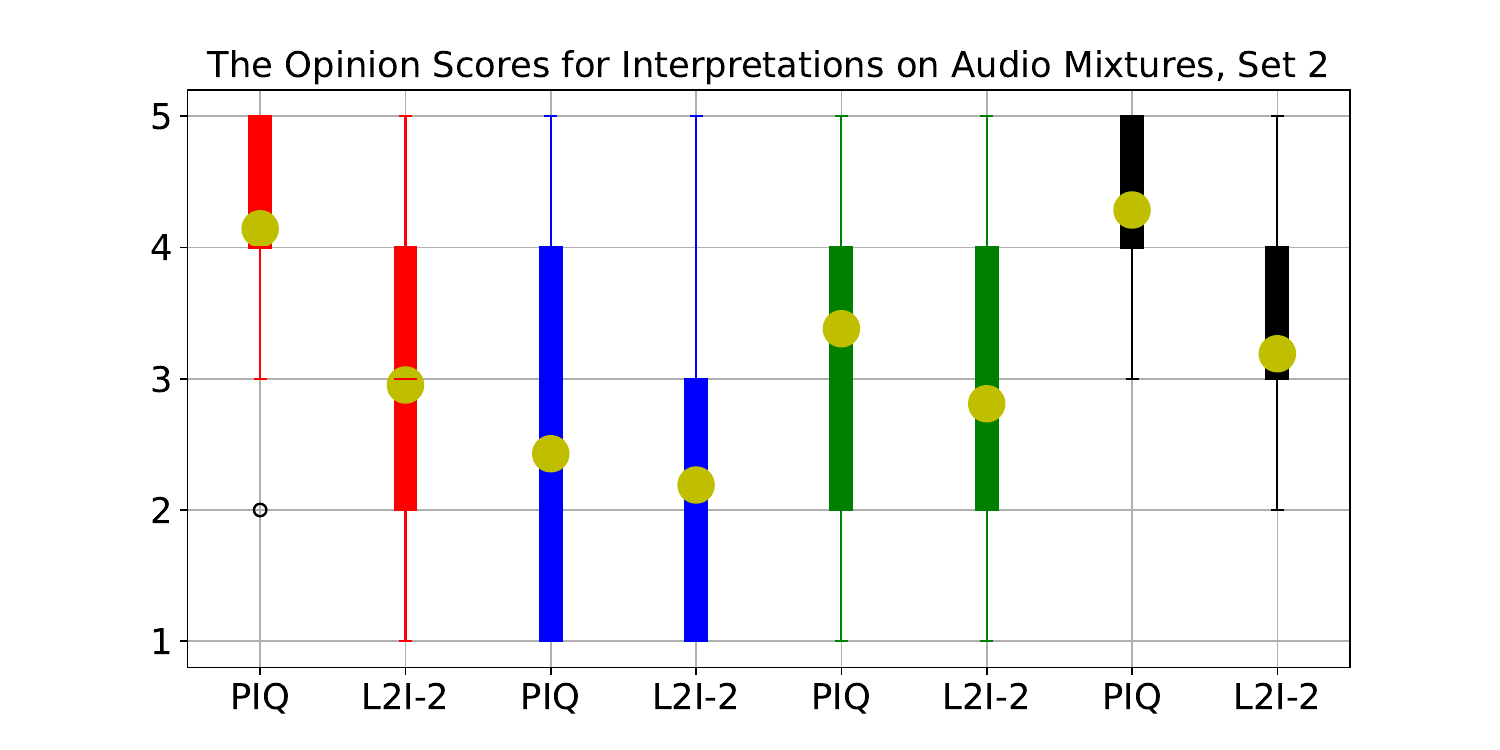}
    \caption{Distribution of user opinion scores on the audio interpretations. Yellow circles indicate the average scores. \textbf{(left)} The distribution of the first set of audio recordings, taken from the official companion website of L2I. \textbf{(right)} Results obtained on audio mixtures we have created. In both sets, we color-coded the audio mixtures: the first recording is red, the second is blue, the third is green, and the fourth is black. The algorithms compared are PIQ (ours), L2I-1 (official results of L2I), L2I-2 (our L2I implementation).}
    \vspace{-.7cm}
    \label{fig:audio1}
\end{figure*}

As we did in Section \ref{sec:images-qualitative} for overlapping images, we examine the interpretation quality of classifier decisions on audio mixtures as well. It is worth recalling that the system is trained on clean signals, not on mixtures. The models we implemented works in the log-magnitude STFT domain, and we reconstruct the time-domain signal by inverting the filtered input magnitude spectrogram using the  phase of the input signal, a common practice in magnitude spectrogram-based source separation, as seen in \cite{deepclustering}.

We compare our method with L2I, as it is recently shown to outperform alternatives for interpreting classifier decisions on audio data \cite{parekh2022listen}. To directly compare the qualitative difference between PIQ and L2I, we tested these methods on the four sound mixtures provided in the \href{https://jayneelparekh.github.io/listen2interpret/}{companion website of L2I}. In addition to these four mixtures, we also tested four different audio mixtures that we created from fold-4 of the ESC50 dataset. To rigorously study the user preference for the interpretations produced by PIQ and L2I, we conducted a user study with 22 participants. On the four sound mixtures provided in the companion website of L2I, we compared PIQ with both i) The official results of L2I from the website (L2I-1) ii) Our implementation of L2I which uses the same classifier as PIQ -- (L2I-2). For the decoder network of L2I-2, we used the same architecture that we used for PIQ, except that we had a pretrained NMF dictionary on the output of the convolutional decoder. We showed the users the mixtures and then asked them to rate the interpretations provided by PIQ, L2I-1, L2I-2 between 1 (bad) and 5 (excellent). We show the result of this user study on the left panel of Figure \ref{fig:audio1}. We see that the participants consistently preferred PIQ over both versions of L2I. These audio interpretations along with the mixtures can be found \href{https://piqinter.github.io/}{on our companion website}.

As previously mentioned, we also compare the interpretation quality of PIQ on four additional mixtures that we created. In this case, we only compared with our implementation of L2I (L2I-2) that interprets the same classifier as PIQ. From the right panel of Figure \ref{fig:audio1}, we can see that users again prefer PIQ over L2I, as shown by the higher average opinion score (represented by yellow circles on top of the box plots).

\vspace{-.4cm}
\section{Conclusions}
\vspace{-.2cm}


In this paper, we proposed PIQ, a post-hoc method for interpreting neural network classifiers. PIQ framework renders it possible to incorporate supervision from foundational models and therefore is able to generate high-quality interpretations for real-life images. Through a series of user studies on image and audio data, we showed that the interpretations generated by PIQ are preferred by participants over several alternative methods in the literature. Furthermore, we demonstrated on black-and-white images that PIQ outperforms several methods on quantitative metrics, and closely matches the original data distribution.

\textbf{Limitations:}
This study is limited to the application of PIQ to image and audio data. In our experiments we have only considered interpretation decoders that generate a fixed size interpretation, but our method does not have a conceptual limitation on this. We have not considered text data, but with a similar use of foundational methods, it is possible to apply PIQ to generate interpretations on text. Note that we discuss the potential societal impacts in Appendix \ref{app:societal}.

\bibliography{neurips_2023}
\bibliographystyle{plainnat}


\newpage
\appendix
\onecolumn

\section{Samples for qualitative evaluation of PIQ on images}
\label{sec:extrapiqsamples}

\begin{figure*}[h!]
   \centering
    \includegraphics[width=0.16\textwidth]{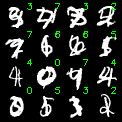}
     \includegraphics[width=0.16\textwidth]{fmnistmixtures/samples.png}
     \includegraphics[width=0.16\textwidth]{fmnistonmnist/samples.png}
     \includegraphics[width=0.16\textwidth]{quickdraw0db/samples.png}
    \includegraphics[width=0.16\textwidth]{quickdraw37/samples.png}
    
    \includegraphics[width=0.16\textwidth]{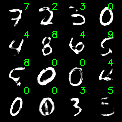}
    \includegraphics[width=0.16\textwidth]{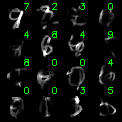}
    \includegraphics[width=0.16\textwidth]{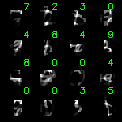}
    \includegraphics[width=0.16\textwidth]{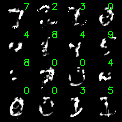}
    \includegraphics[width=0.16\textwidth]{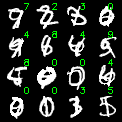}
    \includegraphics[width=0.16\textwidth]{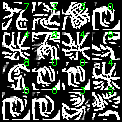}

    \includegraphics[width=0.16\textwidth]{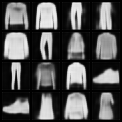}
    \includegraphics[width=0.16\textwidth]{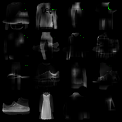}
    \includegraphics[width=0.16\textwidth]{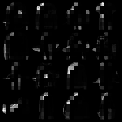}
    \includegraphics[width=0.16\textwidth]{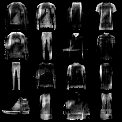}
    \includegraphics[width=0.16\textwidth]{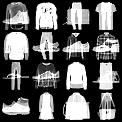}
    \includegraphics[width=0.16\textwidth]{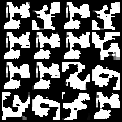}

    \includegraphics[width=0.16\textwidth]{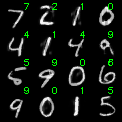}
    \includegraphics[width=0.16\textwidth]{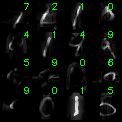}
    \includegraphics[width=0.16\textwidth]{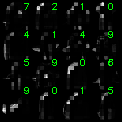}
    \includegraphics[width=0.16\textwidth]{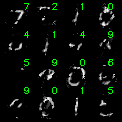}
    \includegraphics[width=0.16\textwidth]{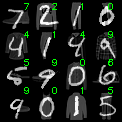}
    \includegraphics[width=0.16\textwidth]{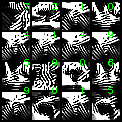}

    \includegraphics[width=0.16\textwidth]{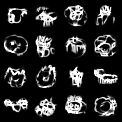}
    \includegraphics[width=0.16\textwidth]{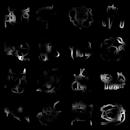}
    \includegraphics[width=0.16\textwidth]{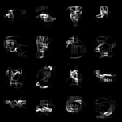}
    \includegraphics[width=0.16\textwidth]{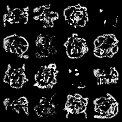}
    \includegraphics[width=0.16\textwidth]{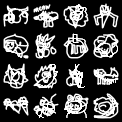}
    \includegraphics[width=0.16\textwidth]{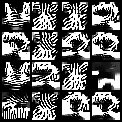}

    \includegraphics[width=0.16\textwidth]{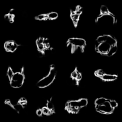}
    \includegraphics[width=0.16\textwidth]{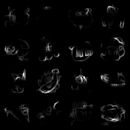}
    \includegraphics[width=0.16\textwidth]{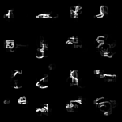}
    \includegraphics[width=0.16\textwidth]{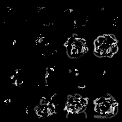}
    \includegraphics[width=0.16\textwidth]{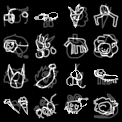}
    \includegraphics[width=0.16\textwidth]{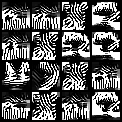}
    
    \resizebox{13.7cm}{!}{
\begin{tabular}{|l|l|l|l|}
\hline
Pullover & Pullover & Trouser & Trouser \\ \hline
Shirt & Trouser & Coat & Shirt \\ \hline
Trouser & Pullover & Coat & Sneaker \\ \hline
Ankle Boot & Dress & Coat & Dress \\ \hline
\end{tabular}
\quad 
\begin{tabular}{|l|l|l|l|}
\hline
Cow & Grapes & Lion & Carrot \\ \hline
Lion & Grapes & Lion & Lion \\ \hline
Lion & Lion & Grapes & Cow \\ \hline
Frog & Grapes & Lion & Lion \\ \hline
\end{tabular}
\quad
\begin{tabular}{|l|l|l|l|}
\hline
Grapes & Ant & Carrot & Frog \\ \hline
Lion & Grapes & Cow & Lion \\ \hline
Dog & Banana & Dog & Lion \\ \hline
Frog & Cow & Cow & Lion \\ \hline
\end{tabular}

}

\caption{The first row shows the input mixtures to the classifier. Comparing interpretation methods on overlapping MNIST digits -  The classifier decisions are indicated on the top right corner of each digit with green text. (second row), overlapping FashionMNIST data items (third row), MNIST digits with FashionMNIST backgrounds (fourth row), overlapping Quickdraw drawings (with equal weights for both images), overlapping Quickdraw drawings (with weights 0.7 and 0.3). The overlapping images that are input to the classifier are shown on the leftmost image. From left-to-right, interpretation images shown correspond to columns 1) PIQ (ours), 2) GradCAM, 3) VIBI, 4) L2I, 5) LIME, 6) FLINT. The table at the bottom of the picture are (from left to right) predicted classes for the second row (case 2 mixtures), predicted classes for the fourth row (case4-i mixtures), and predicted classes for the last row (case4-ii mixtures).}
    \label{fig:overlapmnist_bit}
\end{figure*}

In Figure \ref{fig:overlapmnist}, the interpretations are superimposed on the input samples, Here, we also show the interpretations as grayscale format in Figure \ref{fig:overlapmnist_bit}.

\newpage

\section{L2I adaptation for images}
\label{sec:l2iimpl}

Although L2I \cite{parekh2022listen} was initially presented for audio, the same approach can be applied to images. In particular, Non-Negative Matrix Factorization (NMF) - the core of the L2I approach - has several applications on images \cite{Lee1999LearningTP, Hoyer2004NonnegativeMF}. For the experiments in this paper, we used an NMF dictionary, $W \in \mathbb{R}^{100\times W\times H}$, composed of 100 components of the exact resolution as the original input image ($W\times H$). We kept the architecture of the $\Theta$ network in the original implementation as in the original paper. Thus, it has a pooling layer applied to the spatial dimension, followed by a linear layer, whose weights are used to compute each component's relevance. 
%

\section{Experimental details for Black and White Images}
\label{app:design}
As input for the interpreter, we took the output of the second convolutional block of the classifier, a $4\times4\times128$ tensor. The interpreter decoder consists of transposed convolutional layers (as described in Appendix \ref{app:design}). For all experiments involving black-white images, we used a codebook of 128-dimensional vectors with a total number of 256 vectors. We uniformly divided the dictionary over classes. For the Quickdraw and MNIST datasets, the model output is used to mask the input such that $x_\text{int} = x \odot x_\text{out}$. For FashionMNIST the model output is used as an interpretation directly such that $x_\text{int} = x_\text{out}$. 

For reproducibility, together with the code submitted with this paper we present here, in pseduocode, the main neural networks used in training PIQ for our experiments with images.

The naming convention for the layers is the one from PyTorch\footnote{https://pytorch.org/docs/stable/index.html}. For convolutional layers, $k$ is the kernel size, $s$ is the stride, and $p$ the padding. The classifier architecture is as follows:
\begin{lstlisting}[style=desert, language=PythonCustom]
def classifier_forward(x):
    x = Conv2d(1, 32, k=3, s=1)(x)
    x = ReLU(x)
    x = Conv2d(32, 64, k=3, s=1)(x)
    x = ReLU(x)
    x = MaxPool2d(2, 2)(x)
    x = Dropout2d(p=0.25)(x)
    x = Conv2d(64, 64, k=3, s=1)(x)
    x = ReLU(x)
    x = Conv2d(64, 128, k=3, s=1)(x)
    x = ReLU(x)
    h = MaxPool2d(2, 2)(x) # this is the input for the adapter
    x = Linear(2048, 128)(h)
    x = ReLU(x)
    x = Dropout2d(p=0.5)(x)
    out = Linear(128, num_classes)(x)
    
    return x, h
\end{lstlisting}

The PIQ decoder forward pass is as follows:
\begin{lstlisting}[style=desert, language=PythonCustom]
def decoder_forward(x):
    x = ResBlock(128)(x)
    x = ResBlock(128)(x)
    x = ReLU(x)
    x = ConvTranspose2d(128, 128, k=3, s=2, p=1)(x)
    x = BatchNorm2d(128)(x)
    x = ReLU(x)
    x = ConvTranspose2d(128, 128, k=4, s=2, p=1)(x)
    x = BatchNorm2d(128)(x)
    x = ReLU(x)
    x = ConvTranspose2d(128, 1, k=4, s=2, p=1)(x)
    x = Sigmoid(x)

    return x
\end{lstlisting}
where $\texttt{ResBlock(c)}$ represents a residual block, with an input and output feature map of $c$ channels. The adapter network for PIQ, is a single 3x3 convolutional layer that does not change the number of channels in the feature map.

\section{Experimental details for ImageNet Images}
\label{app:imagenet}
For the subset of the ImageNet dataset, we finetuned a ResNet-50 \cite{He2015DeepRL}, achieving a test accuracy of $88.2\%$. In this case, the interpreter resembles the architecture of a VQ-VAE2 \cite{Razavi2019GeneratingDH}, with class partitioning described in Section \ref{sec:methodology}, applied to the output of the second and last convolutional stage. This way, we can incorporate higher resolution feature maps in the decoding process, while always ensuring the partitioning of the latent space is preserved. 
The two tensors used for reconstructing the interpretation are $32\times32\times512$ and $8\times8\times2048$, respectively. The two codebooks have $4096$ vectors of $2048$ entries each, uniformly distributed over classes. The output of the interpreter is a binary mask that we show on top of the original image (e.g. as in Figure \ref{fig:introshowcase}). 

\begin{figure}[h]
    \centering
    \begin{tikzpicture}[auto, node distance=1.2cm,>=latex']
        \node [] (raw) {\includegraphics[scale=0.4]{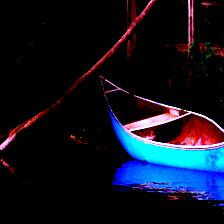}}; 
        \node [above of=raw, scale=0.8, yshift=.7cm] {Input image};
        \node [right of=raw, xshift=3cm] (maskbird) {\includegraphics[scale=0.4]{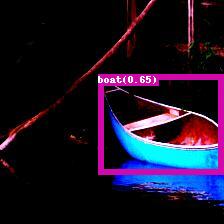}};
        \node [above of=maskbird, scale=0.8, yshift=.7cm] {Target Mask};
        \node [right of=maskbird, xshift=3cm] (sam) {\includegraphics[scale=0.4]{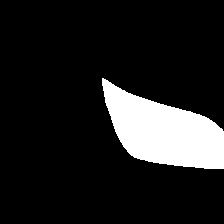}};
        \node [above of=sam, scale=0.8, yshift=.7cm] {SAM mask};


        \draw [->] (raw) -- node [xshift=-0.19cm, yshift=-.5cm, rotate=270, scale=0.7]{G-DINO} (maskbird);
        \draw [->] (maskbird) -- node [xshift=-0.19cm, yshift=-.5cm, rotate=270, scale=0.7]{SAM} (sam);
        
    \end{tikzpicture}

    \caption{Obtaining the training target masks for complex images from the ImageNet dataset. The input image is given to GroundingDINO, which also takes a text prompts and outputs the bounding boxes for all the instances related to the text prompt present in the image. The bounding boxes from GroudingDINO are then used to prompt SAM and generate the target masks.}
    \label{fig:sam}
\end{figure}
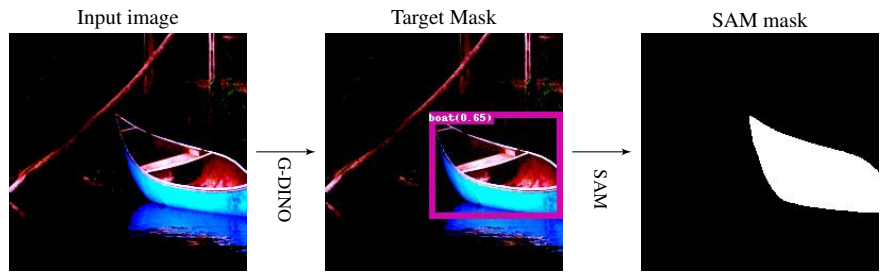

\textbf{Extracting masks with the Segment Anything Model} \\
To extract target masks for colored images, we used the pre-trained image segmentation model SAM \cite{kirillov2023segment}. While the original paper claims that SAM supports text prompting for guiding the segmentation process, this is not supported in the \href{https://github.com/facebookresearch/segment-anything}{official APIs} at the time of writing. Nonetheless, SAM's APIs support prompting with bounding boxes. Thus, we used GroundingDINO\footnote{\href{https://arxiv.org/abs/2303.05499}{https://arxiv.org/abs/2303.05499}} to extract bounding boxes for a specific class and used the output of this process to prompt SAM, as showcased in Figure \ref{fig:sam}. A PyTorch implementation of this process can be found in \href{https://github.com/luca-medeiros/lang-segment-anything}{this GitHub repo}.

We have observed that the prompting SAM with specific ImageNet labels did not yield good quality segmentation masks, and we therefore used coarse labels to prompt SAM and create the segmentation masks (e.g. the class `indigo bunting' gets mapped into the category `bird'). An exaustive mapping for the selected classes in the subset can be found in Table \ref{tab:map}.

\begin{table}[ht]
    \centering
    \caption{Mapping between ImageNet class and coarse label for the selected subset of ImageNet classes.}
    \label{tab:map}
    \begin{tabular}{l|l}
    \textbf{ImageNet Class} & \textbf{Coarse Label} \\
    \hline \hline
    volleyball & ball \\
    ladybug & bug \\
    bathing\_cap & hat \\
    oystercatcher & bird \\
    indigo\_bunting & bird \\
    steel\_drum & instrument \\
    paddle & sports equipment \\
    maillot & clothing \\
    mortarboard & hat \\
    canoe & boat \\
    \end{tabular}
\end{table}

To better explain how we used PIQ in a similar fashion to VQVAE2, we hereafter show the pseudocode for the decoding step of some classifier representations, \texttt{h}.
\begin{lstlisting}[style=desert, language=PythonCustom]
    def decoder_forward(self, hs, labels):
        # hs is a list containing the classifier representations
        h = []
        z_q = []

        # adapter
        hcat = self.conv3(hs[-1])
        hcat = F.normalize(hcat, p=2)
        h.append(hcat)

        # quantize smallest representation
        z_q_x_st = self.codebook(hcat, labels)
        z_q.append(z_q_x_st)
        # upsample quantized small representation
        x_tilde = self.decoder(z_q_x_st) 

        # skip connection with bigger classifier representation
        skip = torch.cat((x_tilde, hs[-3]), dim=1)
        h.append(skip)

        # quantize skip connection output
        z_q_x_st = self.codebook1(skip, labels)
        z_q.append(z_q_x_st)

        # VQVAE2 skip connection and final decoding step
        skip_2 = torch.cat((z_q_x_st, self.upsample_t(hcat)), dim=1)
        x_tilde = self.decoder1(skip_2)

        return x_tilde, h, z_q
\end{lstlisting}

\section{User Study Details for ImageNet}
\label{app:imagenet-users}

\begin{figure}[h]
    \centering
    \includegraphics[scale=0.6]{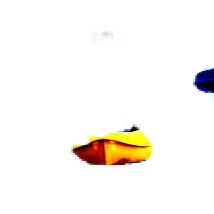}
    \includegraphics[scale=0.6]{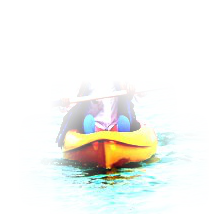}

    \caption{Sample interpretations from the model simulation section of the user study. (\textbf{left}) PIQ interpretations, (\textbf{right}) GradCAM interpretations.}
    \label{fig:context}
\end{figure}

The ImageNet user study consisted of Opinion-Score evaluation and model simulation, as suggested in \cite{Liang2022MultiVizTV}. The results for the Mean-Opinion-Score (MOS) are summarized in Figure \ref{fig:imagenetmos}. Overall, the participants to the user study classified correctly 86.1\% of the PIQ interpretations and 88.0\% of the GradCAM interpretations. However, as shown in Figure \ref{fig:context}, PIQ interpretations are more specific, thus remove more context and make the classification harder for the user. 

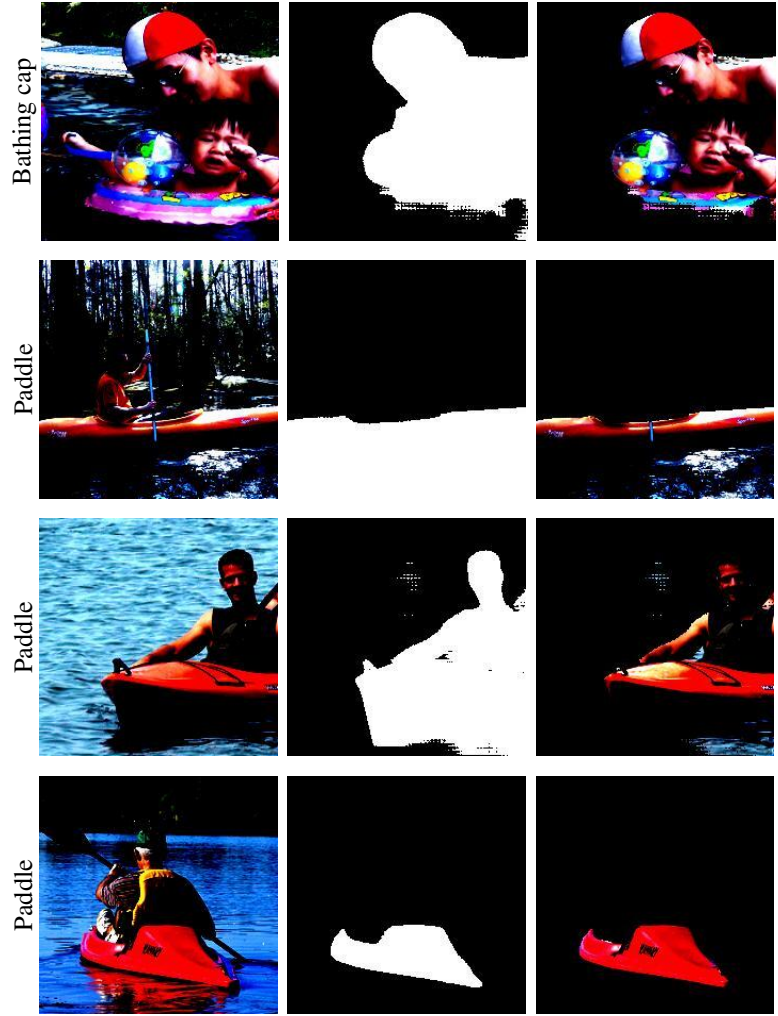
\begin{figure}[h!]
    \centering
    \begin{tikzpicture}[auto, node distance=1.2cm,>=latex']
        \node [] (raw) {\includegraphics[scale=0.4]{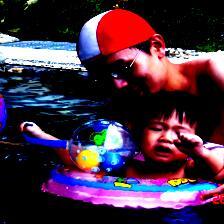}}; 
        \node [left of=raw, xshift=-0.6cm, rotate=90] {Bathing cap};
        \node [right of=raw, xshift=2.1cm] (maskbird) {\includegraphics[scale=0.4]{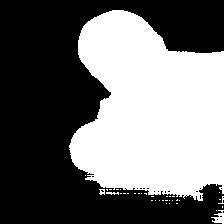}};
        \node [right of=maskbird, xshift=2.1cm] (sam) {\includegraphics[scale=0.4]{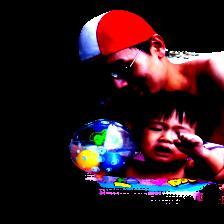}};
    \end{tikzpicture}
    
    \begin{tikzpicture}[auto, node distance=1.2cm,>=latex']
        \node [] (raw) {\includegraphics[scale=0.4]{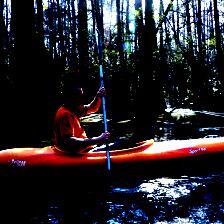}}; 
        \node [left of=raw, xshift=-0.6cm, rotate=90] {Paddle};
        \node [right of=raw, xshift=2.1cm] (maskbird) {\includegraphics[scale=0.4]{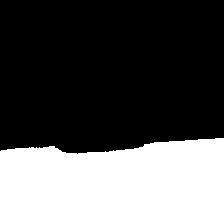}};
        \node [right of=maskbird, xshift=2.1cm] (sam) {\includegraphics[scale=0.4]{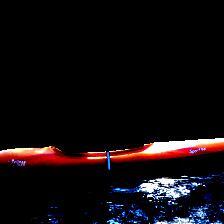}};
    \end{tikzpicture}

    \begin{tikzpicture}[auto, node distance=1.2cm,>=latex']
        \node [] (raw) {\includegraphics[scale=0.4]{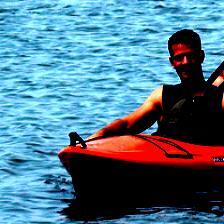}}; 
        \node [left of=raw, xshift=-0.6cm, rotate=90] {Paddle};
        \node [right of=raw, xshift=2.1cm] (maskbird) {\includegraphics[scale=0.4]{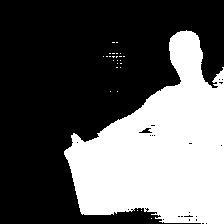}};
        \node [right of=maskbird, xshift=2.1cm] (sam) {\includegraphics[scale=0.4]{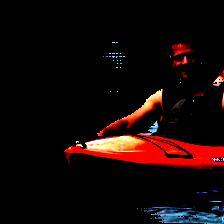}};
    \end{tikzpicture}
    
    \begin{tikzpicture}[auto, node distance=1.2cm,>=latex']
        \node [] (raw) {\includegraphics[scale=0.4]{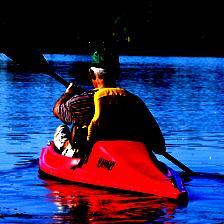}}; 
        \node [left of=raw, xshift=-0.6cm, rotate=90] {Paddle};
        \node [right of=raw, xshift=2.1cm] (maskbird) {\includegraphics[scale=0.4]{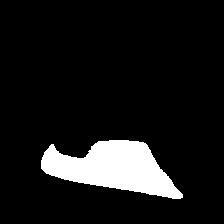}};
        \node [right of=maskbird, xshift=2.1cm] (sam) {\includegraphics[scale=0.4]{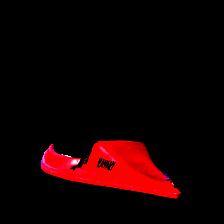}};
    \end{tikzpicture}
    
    \caption{Samples of bad target masks generated using the SAM-GroundingDINO pipeline. From left to right, the images contain the input of the pipeline, the generated segmentation masks (which is used as target by PIQ) and the element-wise multiplication of the two. The first row shows samples from the `bathing cap' class, while the second row shows samples for the `paddle' class.}
    \label{fig:badsam}
\end{figure}

\begin{figure}[b]
    \centering
    \includegraphics[scale=0.6]{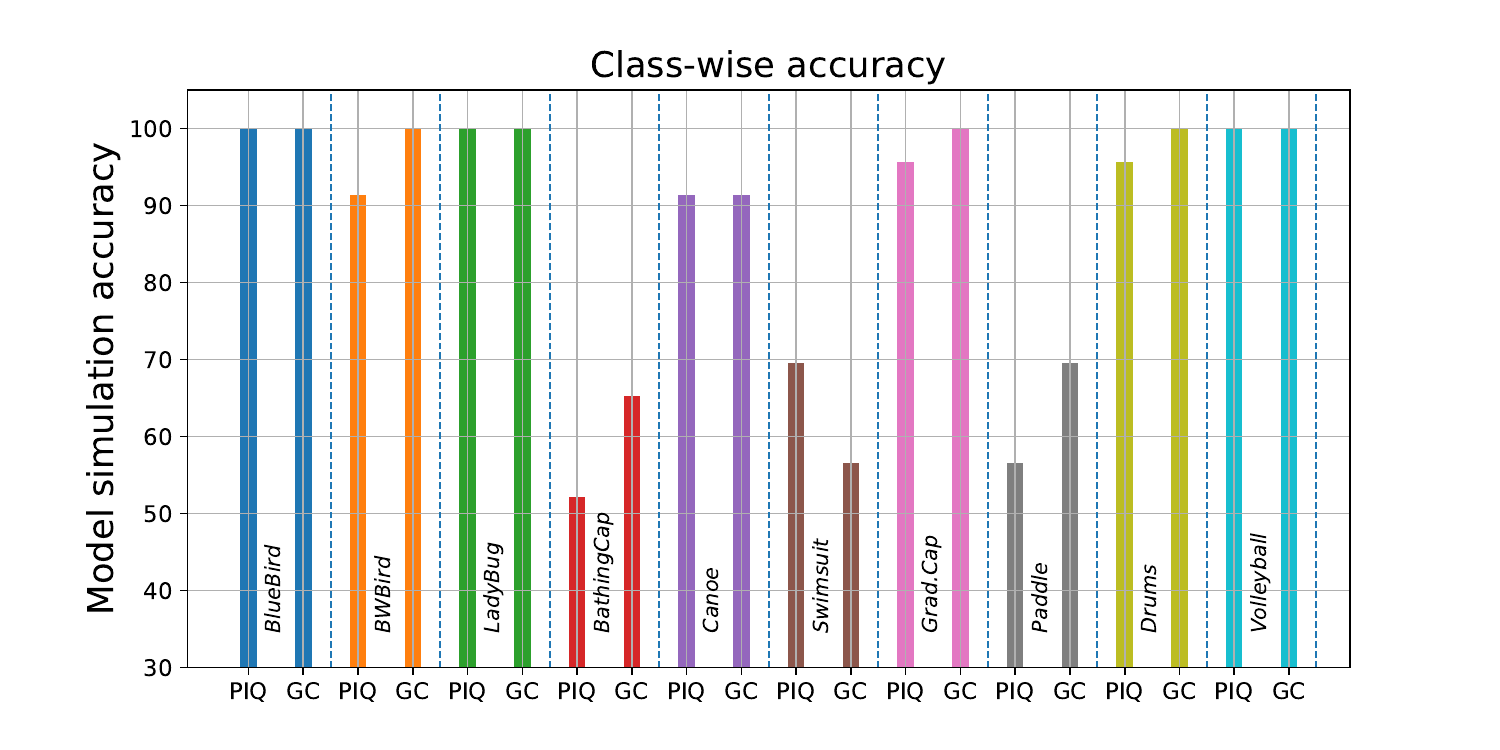}

    \caption{Class-wise model simulation performance for both PIQ and GradCAM.}
    \label{fig:class}
\end{figure}

In Figure \ref{fig:class}, we show the model simulation performance for both PIQ and GradCAM. We observed that among all the classes, `bathing cap' and `paddle', are the hardest to classify for PIQ, when compared to GradCAM. 
To investigate the cause of this drop in interpretation quality, we inspected the target masks given by SAM-GroundingDino pipeline we define in Appendix \ref{app:imagenet}. As shown in Figure \ref{fig:badsam}, we observe that for the classes `paddle' and `bathing cap', the training target masks are not ideal, which potentially diminishes the quality of supervision during training of PIQ. In particular, the `bathing cap' mask example also contains body parts, while the `paddle' examples do not contain paddles at all.



\section{Neural network design for Audio}
\label{app:design_audio}
As we did for the image-oriented implementation of PIQ in Section \ref{app:design}, we present here, with the same notation, the classifier and decoder architectures for the PIQ implementation on audio.

The classifier architecture is as follows:

\begin{lstlisting}[style=desert, language=PythonCustom]
def classifier_forward(x):
    x = Conv2d(1, 256, k=4, s=2, p=1)(x)
    x = BatchNorm2d(256)(x)
    x = ReLU(x)
    
    x = Conv2d(256, 256, k=4, s=2, p=1)(x)
    x = BatchNorm2d(256)(x)
    x = ReLU(x)
    
    x = Conv2d(256, 256, k=4, s=2, p=1)(x)
    x = BatchNorm2d(256)(x)
    x = ReLU(x)
    
    x = Conv2d(256, 256, k=4, s=2, p=1)(x)
    x = BatchNorm2d(256)(x)
    x = ReLU(x)

    h = ResBlock(256)(x)
    x = BatchNorm1d(256)(x)
    
    x = Linear(256, 256)
    x = Linear(256, 50)
    
    return x, h
\end{lstlisting}

The PIQ decoder forward pass is a follows:
\begin{lstlisting}[style=desert, language=PythonCustom]
def decoder_forward(x):
    x = ConvTranspose2d(256, k=256, s=3, p=(2, 2), out_p=1)(x)
    x = ReLU(x)
    x = BatchNorm2d(256)(x)
    x = ConvTranspose2d(256, 256, k=4, s=(2, 2), p=1)(x)
    x = ReLU()(x)
    x = BatchNorm2d(256)(x)
    x = ConvTranspose2d(256, 256, k=4, s=(2, 2), p=1)(x)
    x = ReLU()(x)
    x = BatchNorm2d(256)(x)
    x = ConvTranspose2d(256, 256, k=4, s=(2, 2), p=1)(x)
    x = ReLU()(x)
    x = BatchNorm2d(256)(x)
    x = ConvTranspose2d(256, 1, k=12, s=1, p=1)(x)
    x = Sigmoid(x)

    return x
\end{lstlisting}
where $\texttt{out\_p}$ is the output padding, thus applied to the output of the Conv2dTranspose operation.

\section{Dataset and Modeling Details on Audio}
We test the interpretations produced by PIQ on the ESC-50 dataset \cite{Piczak2015ESCDF}, which consists of 2000, 5 seconds-long clips of 50 different classes of sound events. Example sound events in the dataset include `cat', `dog', `baby cry', `church-bells', and so on.  

As a classifier, we utilized a convolutional network consisting of four strided 2D-convolutional layers with a downsampling factor of 2. Each layer is followed by batch normalization and ReLU activation. The network ends with a residual convolutional layer before a linear classifier. We pretrained the convolutional layers on the VGGSound dataset \cite{Chen20}, which comprises around 550 hours of audio clips sourced from Youtube. The classifier operates in the log-spectrogram domain and achieved 75\% classification accuracy on fold-4 of the ESC50 dataset when trained on folds 1-2-3. We worked with 16kHz audio, using a 1024 point FFT, with a 23ms window-length and 11ms hop length. To balance the distribution of frequencies, we applied a log-transform to the magnitude spectrogram.

 The output of the last layer of the classifier serves as input for the interpreter model. For the adapter, we employed a combination of a residual convolutional layer and a strided 2D convolutional layer. Detailed information on the neural network architectures can be found in Appendix \ref{app:design_audio}. The decoder comprises five layers of strided transposed-2D convolutions. The interpreter is trained on a clean dataset, specifically using folds 1-2-3 of the ESC50 dataset, which is the same dataset used for the classifier. To find a mask on the magnitude STFT, we use PIQ in binary-masking mode and apply a sigmoid nonlinearity at the encoder's output. To obtain the training data for PIQ, we set a threshold of $0.35*\max(X)$ for each spectrogram $X$. We utilized a total of 1024 dictionary items that are evenly distributed across the classes.

 \section{Example Audio Interpretation on ESC50}
\begin{figure}[h!]
    \centering
    \includegraphics[width=.48\textwidth, trim=0cm 0.3cm 0cm 0cm,clip]{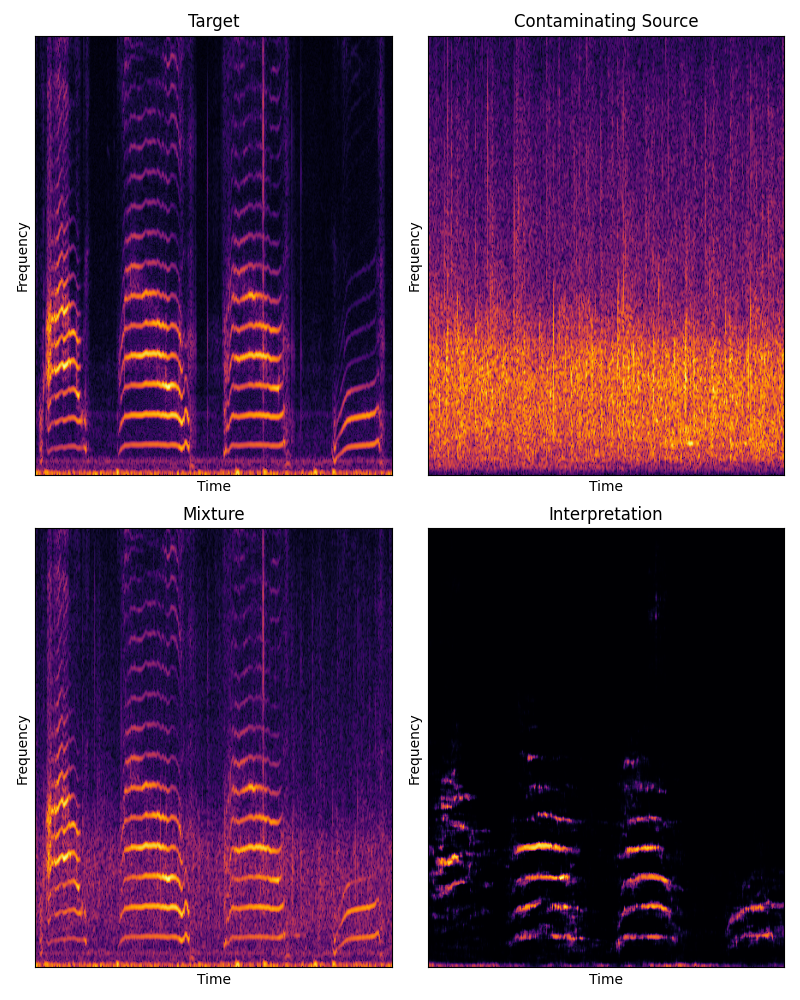}
    \caption{Demonstration of PIQ on audio. (top-left) The dominant audio source, (top-right) Contaminating class, (bottom-left) Mixture, (bottom-right) Produced interpretation.}
    \label{fig:specs}
\end{figure}

 An example of PIQ interpretation on audio can be seen in Figure \ref{fig:specs}. The input signal is a mixture of cat-meowing as the main class and hand clapping as the contaminating class. As shown on the bottom-right spectrogram, the clapping sound is concentrated in the lower half of the spectrum. On the bottom-right panel, we can see that PIQ effectively removes the background clapping noise and focuses on the harmonic of the cat-meowing sound. This interpretation can be found as the 4th mixture in the second section of our companion website\footnote{ \url{https://piqinter.github.io/}}.

\section{Interpretations for Multi-Label Classifiers with PIQ}
\label{app:multilabel}

\begin{figure}[h!]
    \centering
    \includegraphics[width=0.22\textwidth]{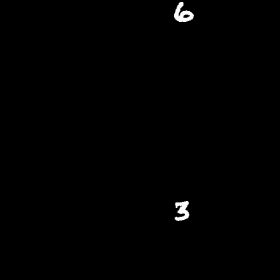}
    \includegraphics[width=0.22\textwidth]{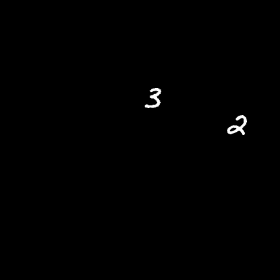}
    \includegraphics[width=0.22\textwidth]{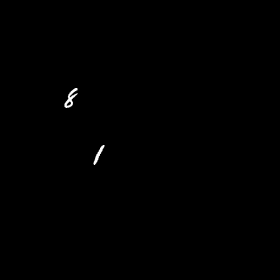}
    \includegraphics[width=0.22\textwidth]{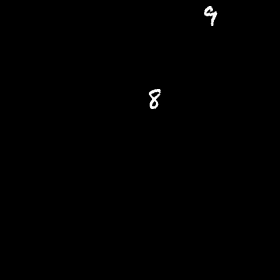}

    \includegraphics[width=0.22\textwidth]{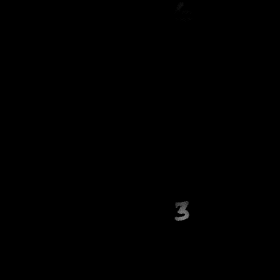}
    \includegraphics[width=0.22\textwidth]{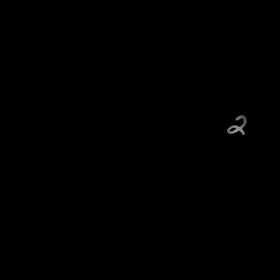}
    \includegraphics[width=0.22\textwidth]{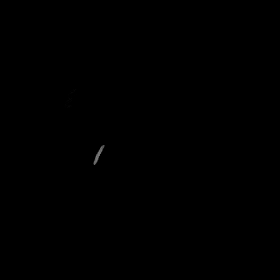}
    \includegraphics[width=0.22\textwidth]{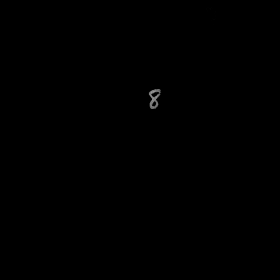}

    \includegraphics[width=0.22\textwidth]{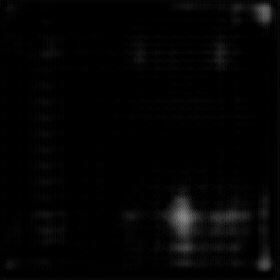}
    \includegraphics[width=0.22\textwidth]{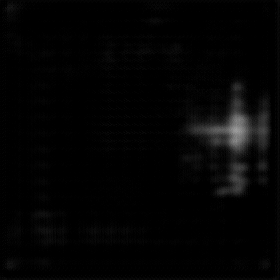}
    \includegraphics[width=0.22\textwidth]{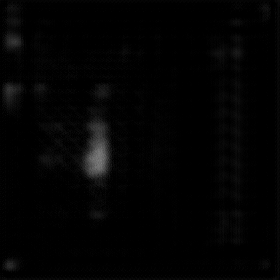}
    \includegraphics[width=0.22\textwidth]{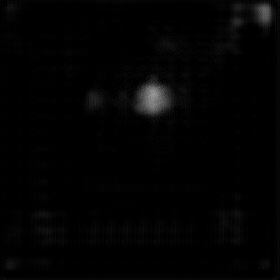}

    \includegraphics[width=0.22\textwidth]{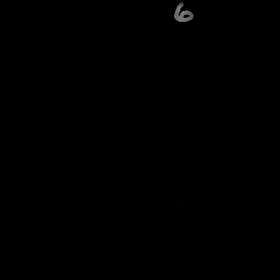}
    \includegraphics[width=0.22\textwidth]{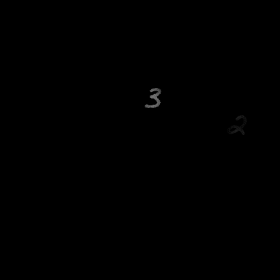}
    \includegraphics[width=0.22\textwidth]{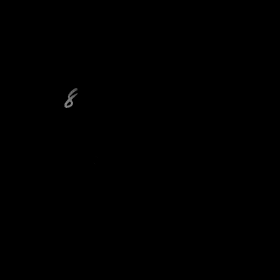}
    \includegraphics[width=0.22\textwidth]{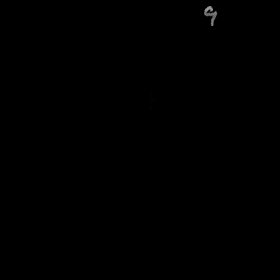}

    \includegraphics[width=0.22\textwidth]{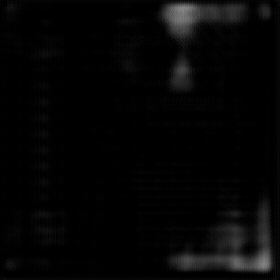}
    \includegraphics[width=0.22\textwidth]{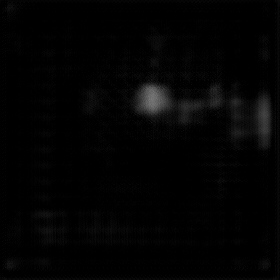}
    \includegraphics[width=0.22\textwidth]{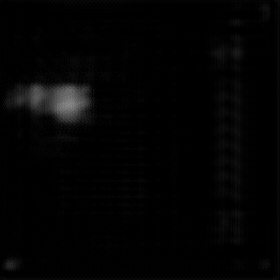}
    \includegraphics[width=0.22\textwidth]{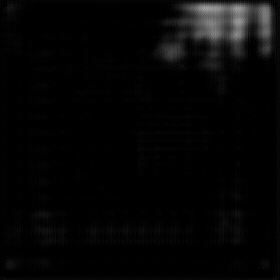}

    \caption{The multi-label classification study where we place the MNIST digits on a larger image. In order, we see images that contain (3,6), (2,3), (1, 8), (8,9). \textbf{(First row)} Input Images. \textbf{(Second Row)} The explanations for the first set of classes \textbf{(Third Row)} The Corresponding Masks for the first set of classes. \textbf{(Fourth Row)} The explanations for the second set of classes. \textbf{(Fifth Row)} The Masks for the second set of classes. }
    \label{fig:scene_mnist}
\end{figure}

We have also explored the possibility of using PIQ in a multilabel classification setting. In order to conduct a preliminary experiment for this, we have created a multilabel classification task where we randomly placed MNIST digits inside an empty image of size $280 \times 280$. We have allowed at most two digits to be active at a time. We show several example images with this dataset, along with the interpretations obtained with PIQ in Figure \ref{fig:scene_mnist}. 

To train PIQ on multi-label data, we adjust the dictionary selection process so that we activate more than one region (as opposed to the multiclass classification case where we only activate one region as we show in Section \ref{sec:methodology}). In addition to the loss function that we have defined in Equation \eqref{eq:trainingloss}, we also add a term that promotes differentiation between the interpretations that correspond to different classes. Overall the loss function we use is as follows: 

\begin{align}
    \mathcal L = &d(x_\text{int}\| x_\text{target}) + \| h' - \text{sg}(h''') \|_2^2 + \| \text{sg}(h') - h''' \|_2^2 + \dots \notag \\ &\; \dots - \sum_{\widehat c \; \in \; E} \Bigg{(} \widehat c^\top \log f(\text{Interpreter}(x_\text{input}\odot x_\text{int}, \widehat c)) + ( 1- \widehat c)^\top \log f(\text{Interpreter}(x_\text{input}\odot x_\text{int}, \widehat c))  \Bigg{)}, 
\end{align}
where $f(\cdot)$ denotes the classifier, and $\text{Interpreter}(\cdot)$ denotes the interpretation model. To the interpretation model, we input the masked image $x_\text{input}\odot x_\text{int}$, and the single class $\widehat c$ which is one of the classes predicted from the classifier for $x_\text{input}$. $E$ is the set of all possible one-hot encoded non-zero vectors, that sum upto the thresholded classifier prediction, which is obtained by thresholding the classifier output $f(x_\text{input})$. Overall, this loss term promotes outputting interpretation masks that only activate one of the classes in the interpretations that correspond to different classes.  

From Figure \ref{fig:scene_mnist}, we observe that even though the digit images appear on the various places in the image, PIQ is able to estimate masks that yield interpretations that highlight the corresponding digits. Note that these results are obtained on a test set (On unseen digits and unseen large images).

\section{Potential Societal Impacts}
\label{app:societal} 

In this work, we propose a method to provide explanations/interpretations for a trained deep neural network. In general, interpretation methods can be used to bolster trust in neural network decisions which can facilitate their use in critical applications such as healthcare. It is possible that bad actors could use this inherently benign technology to create explanations to convince other humans in line with their agenda. We would like to note that we have not worked on mechanisms against this type of misuse, as it is out of scope for this paper.

\section{Computational resources we used in this paper}
All the results presented on this manuscript are obtained using a workstation with two NVIDIA RTX 3090 graphic cards, 64GB of RAM, and an AMD Ryzen 9 7950X CPU. 
\end{document}